\documentclass[11pt]{article}

\usepackage[preprint]{acl}

\usepackage{times}
\usepackage{latexsym}

\usepackage[T1]{fontenc}
\usepackage[utf8]{inputenc}

\usepackage{microtype}

\usepackage{fontawesome}  % 图标支持
\usepackage{graphicx}
\usepackage[table]{xcolor}
\usepackage{makecell}
\usepackage{subcaption}
\usepackage{tabularx}  % 关键：自动调整表格宽度
\usepackage{booktabs}
\usepackage{multirow}
\usepackage{arydshln}
\usepackage{wrapfig}
\title{\begin{minipage}{.06\textwidth}
    \includegraphics[width=\linewidth]{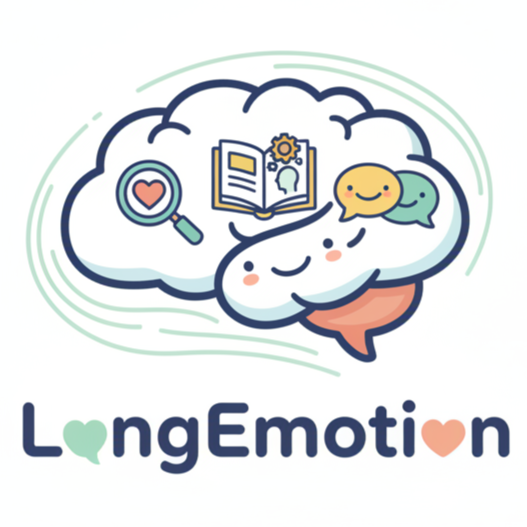} 
    \end{minipage}
    LongEmotion: Measuring Emotional Intelligence of Large Language \\Models in Long-Context Interaction}

\author{
    \linespread{0.92}\selectfont % 调整整体行距
    \setlength{\baselineskip}{11pt} % 控制局部行间距
    \textbf{Weichu Liu\textsuperscript{1}$^{*}$},
    \textbf{Jing Xiong\textsuperscript{2}$^{*}$},
    \textbf{Yuxuan Hu\textsuperscript{3}},
    \textbf{Zixuan Li\textsuperscript{4}},
    \textbf{Minghuan Tan\textsuperscript{5}},
    \textbf{Ningning Mao\textsuperscript{6}},
    \textbf{Hui Shen\textsuperscript{2,7}},\\[3pt]
    \textbf{Wendong Xu\textsuperscript{2}},
    \textbf{Chaofan Tao\textsuperscript{2}},
    \textbf{Min Yang\textsuperscript{5}$^{\dagger}$},
    \textbf{Chengming Li\textsuperscript{1}$^{\dagger}$},
    \textbf{Lingpeng Kong\textsuperscript{2}},
    \textbf{Ngai Wong\textsuperscript{2}}\\[8pt]
    \textsuperscript{\rm 1}Shenzhen MSU-BIT University,
    \textsuperscript{\rm 2}The University of Hong Kong,
    \textsuperscript{\rm 3}City University of Hong Kong,\\[-1pt]
    \textsuperscript{\rm 4}Institute of Automation, Chinese Academy of Sciences,\\[-1pt]
    \textsuperscript{\rm 5}Shenzhen Institute of Advanced Technology, Chinese Academy of Sciences,\\[-1pt]
    \textsuperscript{\rm 6}Beijing Normal University,
    \textsuperscript{\rm 7}University of Michigan, Ann Arbor\\[6pt]
    \small{
        \faGithub~\textbf{Github:} \href{https://longemotion.github.io/}{https://longemotion.github.io/}\quad
        \raisebox{-0.5mm}{\includegraphics[height=3.5mm]{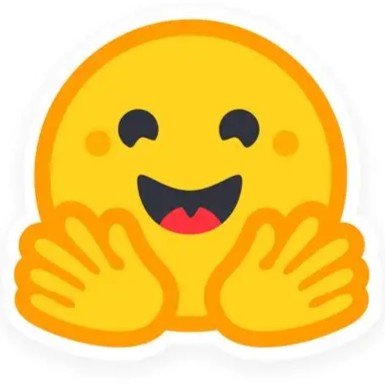}}~\textbf{Huggingface:} \href{https://huggingface.co/datasets/LongEmotion/LongEmotion}{LongEmotion}
    }
}

\begin{document}
\maketitle

\renewcommand{\thefootnote}{\fnsymbol{footnote}}
\setcounter{footnote}{0}
\footnotetext[1]{Equal contribution.}
\footnotetext[2]{Corresponding author.}

\begin{abstract}
Large language models (LLMs) make significant progress in Emotional Intelligence (EI) and long-context modeling. However, existing benchmarks often overlook the fact that \textit{emotional information processing unfolds as a continuous long-context process}. To address the absence of multidimensional EI evaluation in long-context inference and explore model performance under more challenging conditions, we present \textsc{LongEmotion}, a benchmark that encompasses a diverse suite of tasks targeting the assessment of models’ capabilities in \textbf{Emotion Recognition}, \textbf{Knowledge Application}, and \textbf{Empathetic Generation}, with an average context length of 15,341 tokens. To enhance performance under realistic constraints, we introduce the Collaborative Emotional Modeling (\textsc{CoEM}) framework, which integrates Retrieval-Augmented Generation (\textsc{RAG}) and multi-agent collaboration to improve models’ EI in long-context scenarios. We conduct a detailed analysis of various models in long-context settings, investigating how reasoning mode activation, RAG-based retrieval strategies, and context-length adaptability influence their EI performance.
\end{abstract}

%  Furthermore, we conduct a detailed case study on the performance comparison among GPT series models, the application of CoEM in each stage and its impact on task scores, and the advantages of the LongEmotion dataset in advancing EI.

\section{Introduction}
Large Language Models (LLMs) are increasingly adopted in the domain of Emotional Intelligence (EI)~\citep{wang2023emotional}. By leveraging their advanced language understanding and generation capabilities, LLMs become valuable tools for facilitating emotional expression~\citep{ishikawa2025ai,lu2025understanding}, with recent work showing their capacity to simulate specified emotional states in accordance with established models such as Russell’s Circumplex~\citep{russell1980circumplex,russell2003core}. LLMs are increasingly serving in roles ranging from mental health assistants~\citep{guo2403large,malgaroli2025large,fu2024laerc} to everyday conversational companions~\citep{fu2024laerc,duanexploration,zhang2025cdea}. This growing integration into emotionally sensitive domains places greater demand on LLMs to maintain emotional coherence over time — not only to understand but also to remember, adapt, and respond empathetically in prolonged inference~\citep{zhong2024memorybank}.

Although existing benchmarks make considerable progress in measuring the EI of LLMs~\citep{sabour2024emobench, emotion-bench}, current evaluation still suffers from the following limitations:
(i) As articulated by Affective Information Processing Theory~\citep{lang1984affective}, \emph{humans continuously receive, process, organize, and respond to emotional information}, which can manifest unique patterns of emotional intelligence within a long-context setting. Existing studies often overlook the gap between idealized conditions and real-world scenarios: in realistic settings, the processing of emotional information is a continuous and enduring process. To bridge this gap, models should be evaluated on their EI in long context, which can be further decomposed into three key abilities: \textit{accurate emotion recognition, appropriate knowledge application, and affectively empathetic expression in long-context inference.}  (ii) Current research predominantly focuses on measuring a single aspect of the model's capabilities, such as classification, expression, etc. According to the Mayer-Salovey-Caruso Emotional Intelligence Test (MSCEIT)~\citep{mayer2002mayer}, an individual's EI encompasses multiple dimensions. Assessing only one specific capability is insufficient to fully represent the model's EI. (iii) The emotional behavior of recent state-of-the-art techniques in long-context scenarios remains unexplored, especially for reasoning models with think mode, RAG-based agent methods, and other emerging approaches.

\begin{figure*}[htbp]
  \centering
  % 下方的两个图（子图b）
  \subcaptionbox{Token distributions across tasks.\label{fig:subfig2}}{%
    \includegraphics[width=0.47\textwidth]{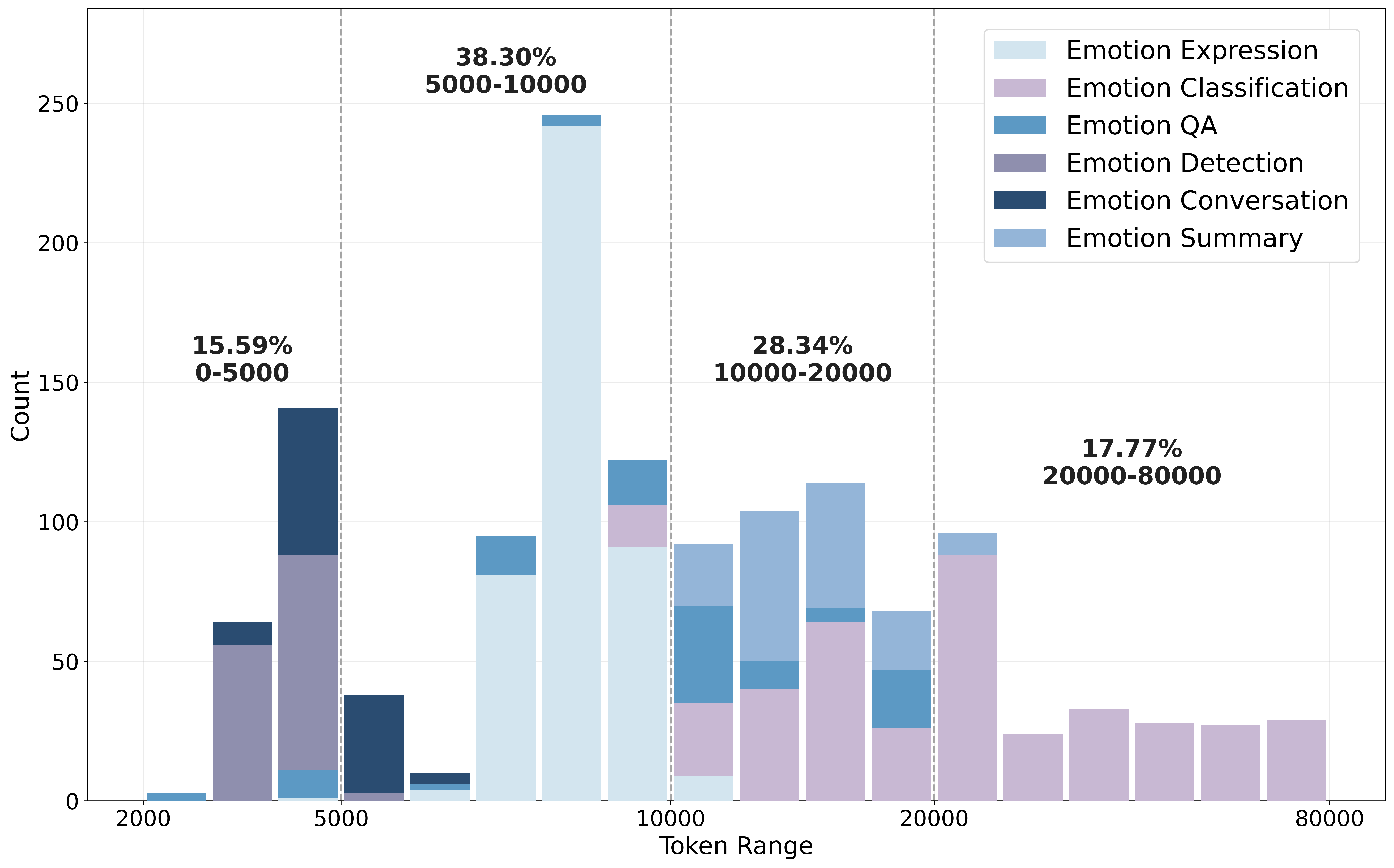}
  }
  \hspace{0.03\textwidth} % 使用更小的间距
  \subcaptionbox{Distribution of sample counts.\label{fig:subfig3}}{%
    \includegraphics[width=0.45\textwidth]{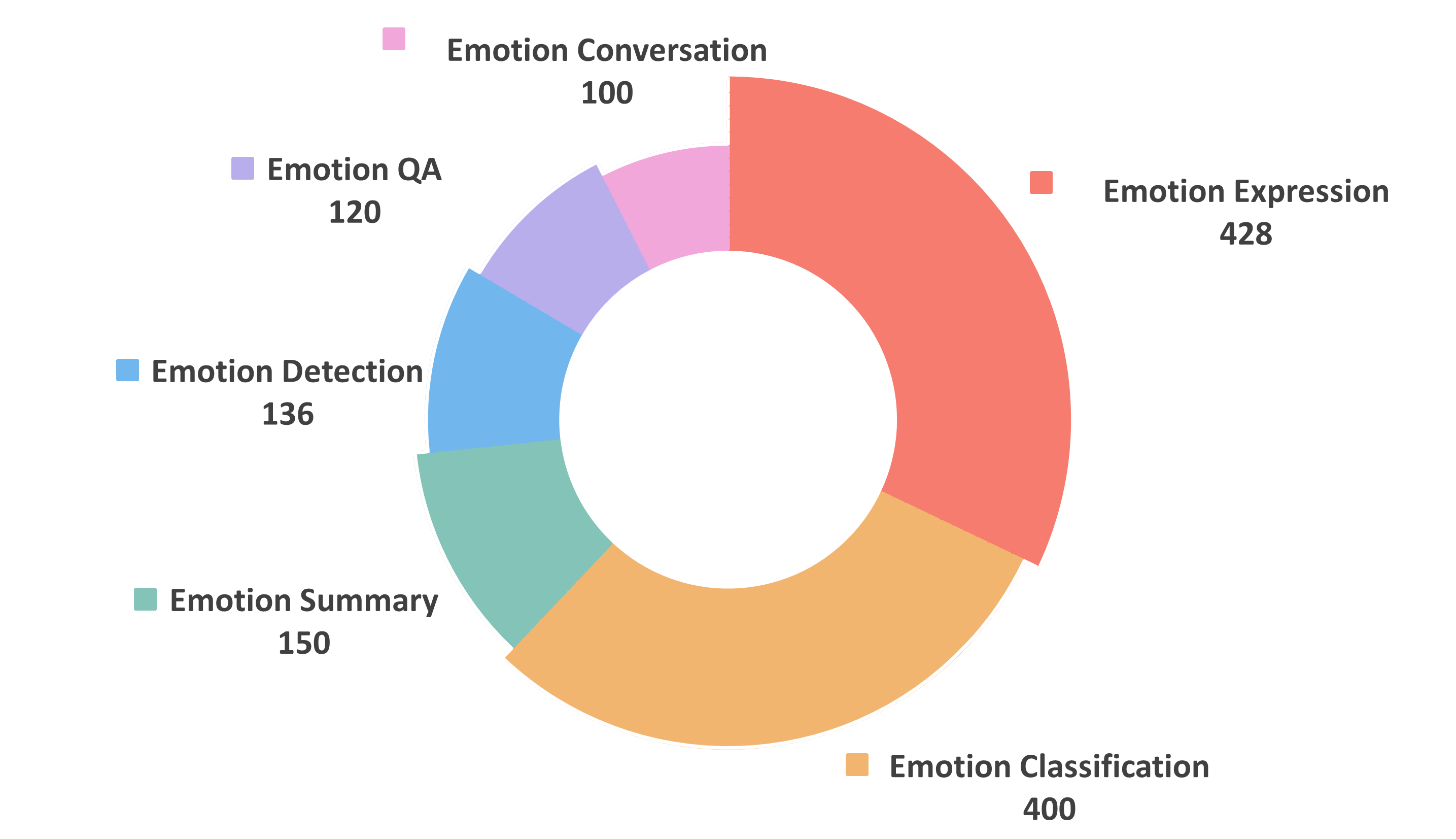}
  }
  
  \caption{(a) Sequence length denotes average model output length for Emotion Expression, and average input context length for other tasks. (b) Distribution of sample counts across the six tasks, illustrating the overall composition of the dataset.}
  \label{fig1}
\end{figure*}

To bridge realistic scenarios and long-context evaluation, we introduce \textsc{LongEmotion}, a benchmark designed to comprehensively evaluate the EI of LLMs in long-context inference. \textsc{LongEmotion} comprises six complementary tasks. Two \textit{Emotion Recognition} tasks, Emotion Classification and Emotion Detection, measure the model's reasoning ability when key emotional information is located in noisy, long-context scenarios; two \textit{Empathetic Generation} tasks, Emotion Conversation and Emotion Expression, evaluate the model's empathy and expression abilities in the context of expansive multi-turn conversations or self-narratives; two \textit{Knowledge Application} tasks, Emotion QA and Emotion Summary, probe how effectively the model leverages and applies emotional knowledge in authentic scenarios. Figure~\ref{fig1} depicts the dataset’s distribution.

To handle these realistic settings, we develop a Retrieval‐Augmented Generation (RAG) approach as well as a novel multi‐agent emotional modeling framework called Collaborative Emotional Modeling (\textsc{CoEM}). Unlike standard RAG systems that pull from static, external corpora, our method treats the conversation history itself as a dynamic vector store to capture aspect-level sentiment terms. To further enhance EI in long context, we introduce \textsc{CoEM}, where the context is divided into coherent chunks, initially ranked by relevance, and then processed by multiple collaborating agents (e.g., an auxiliary GPT-4o instance~\citep{openai-gpt-4o}). After a second‐stage re‐ranking, these agents collectively generate an emotional “ensemble” response. 
This architecture captures the uncertainty and fluidity of real-world dialogue, allowing emotionally salient information to be continuously extracted, re-contextualized, and articulated.
To further investigate the applicability of RAG techniques to long-context-based emotional tasks, we also adapt Self-RAG~\citep{asai2024self} and Search-o1~\citep{li2025search} methods to LongEmotion by replacing their retrieval corpus with conversational context, thereby exploring broader possibilities of RAG in the domain of Emotional Intelligence. Our contributions are summarized as:
\begin{itemize}
\item We present \textsc{LongEmotion}, a long-context EI benchmark with six diverse tasks targeting recognition, generation, and psychological knowledge application.

\item We propose \textit{CoEM} framework to enhance performance by retrieving and enriching contextually relevant information.

\item We perform extensive experiments across all settings and comprehensive case study, offering detailed analyses of LLMs' EI in long-context scenarios.
\end{itemize}
\section{Related Work}
\label{sec:related-work}

\begin{figure*}[t]
\centering
    \includegraphics[width=1\textwidth]{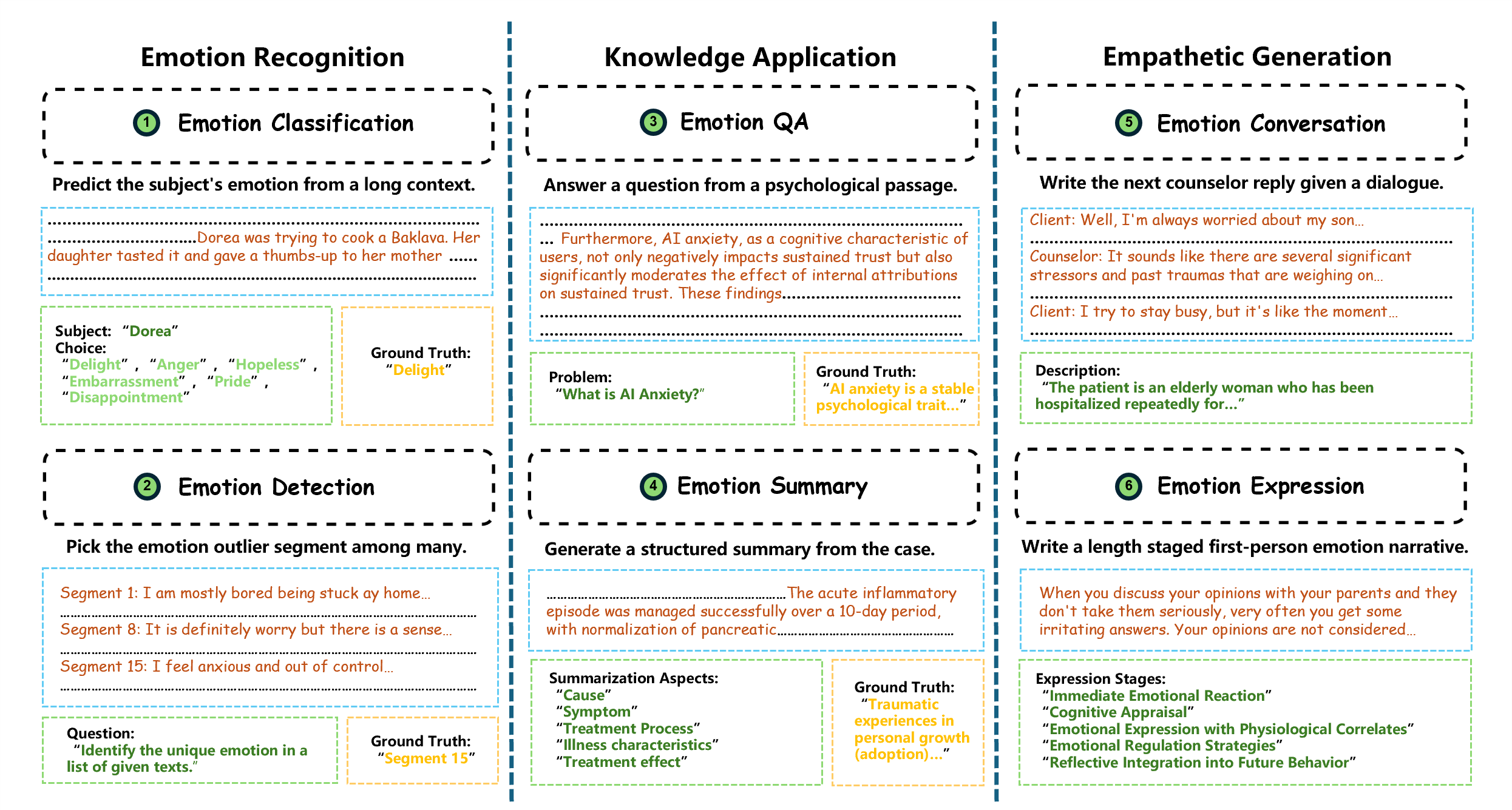} % 稍微小于0.5避免溢出
    \caption{An illustrative overview of the LongEmotion dataset. To comprehensively evaluate the EI of LLMs in long-context interaction, we design six tasks: Emotion Classification, Emotion Detection, Emotion QA, Emotion Conversation, Emotion Summary, and Emotion Expression. }
    \label{fig:overview_of_longemotion_dataset}
\end{figure*}
\paragraph{Emotional Intelligence Benchmarks.}

Many benchmarks are developed to assess LLMs’ Emotional Intelligence (EI). Emobench~\citep{sabour2024emobench} draws on psychological theories to evaluate both emotional understanding and application across 400 English–Chinese handcrafted questions, exposing significant gaps between model and human EI levels. EQ‑Bench~\citep{paech2023eq} measures LLMs' ability to rate emotional intensity in dialogues through 60 English queries, showing strong correlation with multi-domain reasoning benchmarks. More recently, EmotionQueen~\citep{chen2024emotionqueen} offers a specialized benchmark for empathy, requiring LLMs to recognize key events, implicit emotions, and generate empathetic responses. Despite their strengths, all of these focus on short or synthetic interactions and lack the long contextual depth critical for assessing EI in extended conversational or narrative settings.

\paragraph{Long-Context Modeling.}

LLMs make strides in processing long documents, yet robust evaluation remains an open challenge. LongBench~\citep{bai2023longbench} introduces a bilingual, multi-task benchmark covering QA, summarization, and code tasks with average context lengths over 6,000 words, revealing that even state-of-the-art models struggle with extended inputs. Complementing this, LooGLE~\citep{li2023loogle} evaluates long-context reasoning using realistic documents exceeding 24k tokens, uncovering dependencies that span across distant spans. For extreme-length evaluation, XL$^2$Bench~\citep{ni2024xl} includes tasks on fiction, law, and scientific papers with inputs up to 100k+ words—yet LLMs still fall short in handling long-range dependencies. Beyond these, RULER~\citep{chen2023ruler} focuses on complex reasoning chains in long-form texts via fine-grained question types and inter-paragraph dependencies, providing a valuable diagnostic lens into model reasoning depth. InfiniteBench~\citep{sun2024infinitebench}, meanwhile, evaluates LLMs’ abilities on open-ended, unbounded contexts with theoretically unlimited input lengths, highlighting model degradation as input exceeds trained context windows. Survey work such as \citet{liu2025comprehensive} offers a broad overview of long-context modeling and evaluation paradigms but emphasizes that most benchmarks primarily target information retrieval or general comprehension—not emotional intelligence or affective computing.
\section{LongEmotion: Construction and Task}
\label{sec:construction-and-task}
A visual overview is shown in Figure~\ref{fig:overview_of_longemotion_dataset}. Appendix \ref{llmjudge-Metrics} provides a detailed explanation of metrics used in tasks where LLMs act as evaluators. We summarize the advantages of LongEmotion in enhancing LLMs' EI in Appendix \ref{appendix:advantage-of-longemotion}
\begin{table*}[!t]
\centering
\small % 字体与正文协调、略小
\setlength{\tabcolsep}{6pt} % 可微调列间距
\renewcommand{\arraystretch}{1.0} % 控制行距

\begin{tabularx}{0.95\textwidth}{
    l
    c
    l
    l
    c
    c
    c
}
\toprule
\textbf{Task} & \textbf{ID} & \textbf{Source} & \textbf{Construction} & \textbf{Metric} & \textbf{Avg len} & \textbf{Count} \\

\multicolumn{7}{c}{\cellcolor[HTML]{E5E5FC}\textit{\textbf{Emotion Recognition}}} \\
Emotion Classification & EC & Emobench, FinEntity & Segment Insertion & Accuracy & 30139 & 400 \\
Emotion Detection & ED & Covid-worry & Reorganization & Accuracy & 4106 & 136 \\

\multicolumn{7}{c}{\cellcolor[HTML]{E5E5FC}\textit{\textbf{Knowledge Application}}} \\
Emotion QA & QA & Literature & Human Annotation & F1 & 11207 & 120 \\
Emotion Summary & ES & CPsycoun & Human Annotation & LLM as Judge & 15341 & 150 \\

\multicolumn{7}{c}{\cellcolor[HTML]{E5E5FC}\textit{\textbf{Empathetic Generation}}} \\
Emotion Conversation & MC & CPsycoun & Expansion & LLM as Judge & 4856 & 100 \\
Emotion Expression & EE & EmotionBench & Reorganization & LLM as Judge & 8546* & 428 \\
\bottomrule
\end{tabularx}

\caption{
A statistical overview of the \textsc{LongEmotion} dataset. 
\textit{ID} denotes task abbreviations. EC, ED, QA, MC, and ES involve long-text input, with \textit{Avg len} showing average context length. 
EE is a long-text generation task—\textit{Avg len} here refers to average output length (marked with *).}
\label{dataset-overview}
\end{table*}

\subsection{Task Design}
\paragraph{Emotion Classification.}
This task requires the model to identify the emotional category of a target entity within long-context texts that contain lengthy spans of context-independent noise~\citep{NIAH}. Model performance is evaluated by its accuracy against the ground truth.

\paragraph{Emotion Detection.}
The model is given N+1 emotional segments. Among them, N segments express the same emotion, while one segment expresses a unique emotion. The model is required to identify the single distinctive emotional segment. During evaluation, the model's score depends on whether the predicted index matches the ground-truth index.

\paragraph{Emotion QA.}
In this task, the model is required to answer questions grounded in long-context psychological literature. Model performance is evaluated using the F1 score between its responses and the ground truth answers.

\paragraph{Emotion Summary.}
In this task, the model is required to summarize the following aspects from long-context psychological pathology reports: (i) causes, (ii) symptoms, (iii) treatment process, (iv) illness characteristics, and (v) treatment effects. After generating the model’s response, we employ GPT-4o to evaluate its factual consistency, completeness, and clarity with respect to the reference answer. These three evaluation criteria are validated in CPsyExam~\citep{zhao2024cpsyexam}.

\paragraph{Emotion Conversation.}
In our four-stage long-context counseling dialogue dataset, we select the quartile, half, and three-quarter points of each stage as evaluation checkpoints to assess the model's EI capabilities. We introduce 12 specialized metrics informed by five major therapeutic frameworks: Cognitive Behavioral Therapy (CBT)~\citep{beck2021cbt}, Acceptance and Commitment Therapy (ACT)~\citep{waltz2010act}, Humanistic Therapy~\citep{humanisitic}, Existential Therapy~\citep{existential-psychology}, and Satir Family Therapy~\citep{satir-family-therapy}, which can be seen in Appendix \ref{llmjudge-Metrics}. The scoring is performed by GPT-4o, which serves as the evaluator to ensure consistency and scalability.

\paragraph{Emotion Expression.}
In this task, the model is situated within a specific emotional context and prompted to produce a long-form emotional self-narrative. Models first complete a psychometric self-assessment (e.g., PANAS), followed by the generation of a structured narrative spanning five phases: (i) Immediate Reaction, (ii) Cognitive Appraisal, (iii) Emotional and Physiological Expression, (iv) Regulation Strategies, and (v) Reflective Integration. The evaluation encompasses six dimensions: emotional consistency, content redundancy, expressive richness, cognition–emotion interplay, self-reflectiveness, and narrative coherence. All dimensions are assessed by GPT-4o, which serves as the evaluator to score the model’s capacity for emotional expression.

\subsection{Data Construction}
The statistical overview of LongEmotion dataset can be found in Table~\ref{dataset-overview}. EC and ED tasks focus on evaluating the model’s ability in emotional recognition. QA and ES tasks emphasize the model’s capability to apply knowledge within long-context scenarios. MC and EE tasks aim to measure the model’s generative ability.

\paragraph{Reorganization from Existing Datasets.}
In Emotion Classification, we embed short excerpts from Emobench~\citep{sabour2024emobench} and FinEntity~\citep{tang2023finentity} into BookCorpus passages~\citep{Zhu_2015_ICCV}, by randomly inserting snippets and manually adjusting proper nouns for coherence. In Emotion Detection, we build contrast sets by grouping texts from Covid-worry~\citep{covid-worry,van2023multi} by emotion label and inserting mismatched segments. In Emotion Expression, we use \textit{situations} from EmotionBench~\citep{emotion-bench} to provide models with specific emotional contexts.

\paragraph{Expansion and Human Annotation}

\begin{figure}[htbp]
    \centering
    \includegraphics[width=0.45\textwidth]{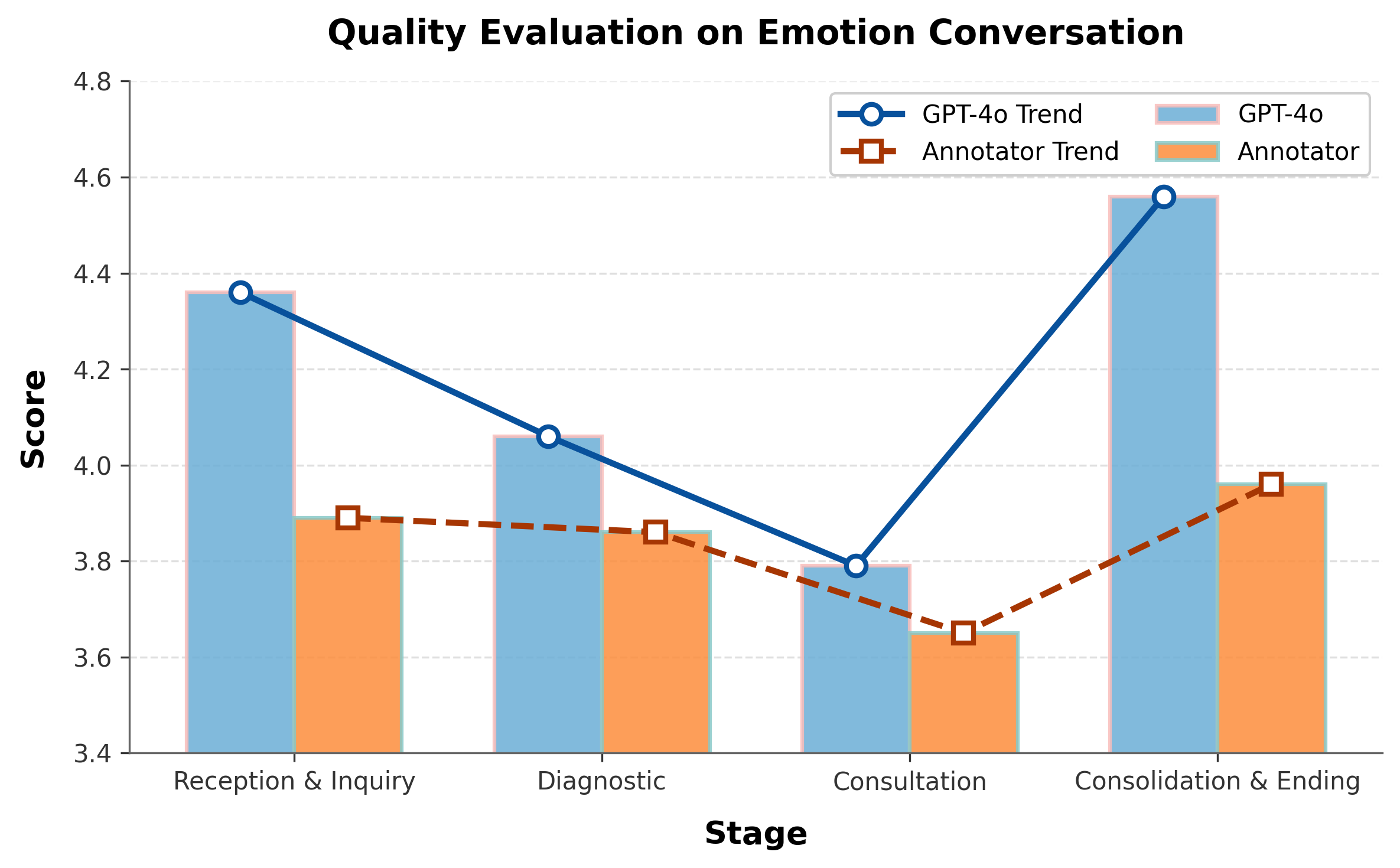} % 稍微小于0.5避免溢出
    \caption{Quality Evaluation on Emotion Conversation.}
    \label{fig:multi_turn_assessment}
\end{figure}

For Emotion Conversation, based on CPsyCoun~\citep{zhang2024cpsycoun}, we construct 100 emotionally rich dialogues by expanding seed prompts into four functional stages: (i) Reception and Inquiry, (ii) Diagnostic, (iii) Consultation, and (iv) Consolidation and Ending. Dataset quality is evaluated through two parallel protocols: (i) manual scoring by psychology experts and (ii) automated assessment with GPT-4o. As reported in Figure~\ref{fig:multi_turn_assessment}, the Pearson correlation between LLM and human scores reaches 0.934 ($p = 0.066$), indicating a relatively high alignment. In addition, we use the same prompts and GPT model for evaluation as those employed in the quality assessment, which further validates the rationality of our LLM-as-Judge setting. Annotator qualifications are detailed in Appendix~\ref{Qualifications-of-annotators}.

In Emotion Summary, drawing on CPsyCounR dataset, we first expand the \textit{experience and reflection} section of the dataset to meet our requirements for long-context inputs. Next, psychology annotators label each sample across five standardized dimensions: (i) Causes, (ii) Symptoms, (iii) Treatment Process, (iv) Illness Characteristics, and (v) Treatment Effect. Finally, by filtering samples based on format, content richness, and precision, we select a final set of 150 samples. To further extend the dataset length while preserving the original semantic integrity, we employ DeepSeek-V3~\citep{deepseekai2024deepseekv3technicalreport} to perform structured decomposition and subsequent content augmentation. In Appendix \ref{appendix:advantage-of-longemotion}, we discuss the annotation discipline for the annotation process of Emotion Summary.

\begin{figure}[htbp]
    \centering
    \includegraphics[width=0.38\textwidth]{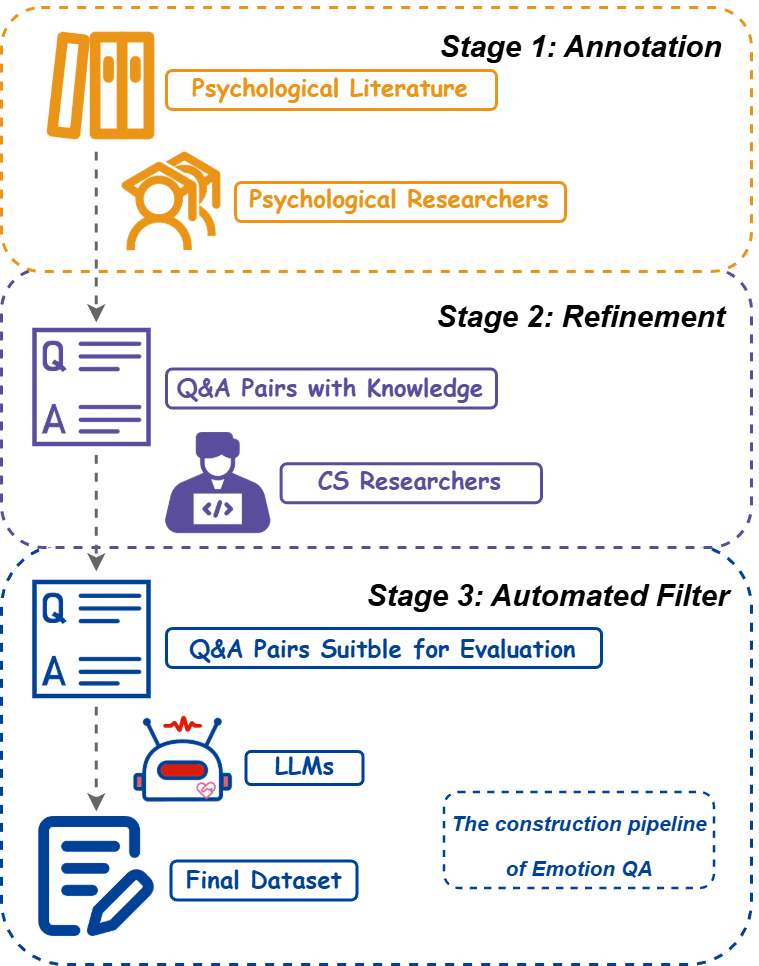} % 稍微小于0.5避免溢出
    \caption{Annotation process of Emotion QA.}
    \label{fig:data-annotation}
\end{figure}

In constructing Emotion QA, the annotation pipeline is illustrated in Figure \ref{fig:data-annotation}. The construction process on psychological literature involves: (i) expert-written questions targeting emotional understanding, (ii) refinement of reference answers for clarity and consistency with F1-based evaluation, and (iii) filtering based on model performance to exclude overly ambiguous or trivial examples. Through this series of manual annotation and selection, we finally obtain 120 high-quality pairs of psychological knowledge questions and answers.

\section{Collaborative Emotional Modeling}
\label{sec:method}
Figure \ref{fig:pipeline} illustrates the pipeline of CoEM. To address EI tasks involving long contexts, we propose a hybrid retrieval-generation architecture that combines Retrieval-Augmented Generation (RAG) with modular multi-agent collaboration. For the parameter settings and application details, please refer to Appendix \ref{appendix:Details-rag-coem}. For the case analysis of RAG and CoEM, please refer to Appendix \ref{case-analysis-of-rag-coem}. The framework consists of five key stages:
\begin{figure*}[htbp]
\centering
    \includegraphics[width=0.90\textwidth]{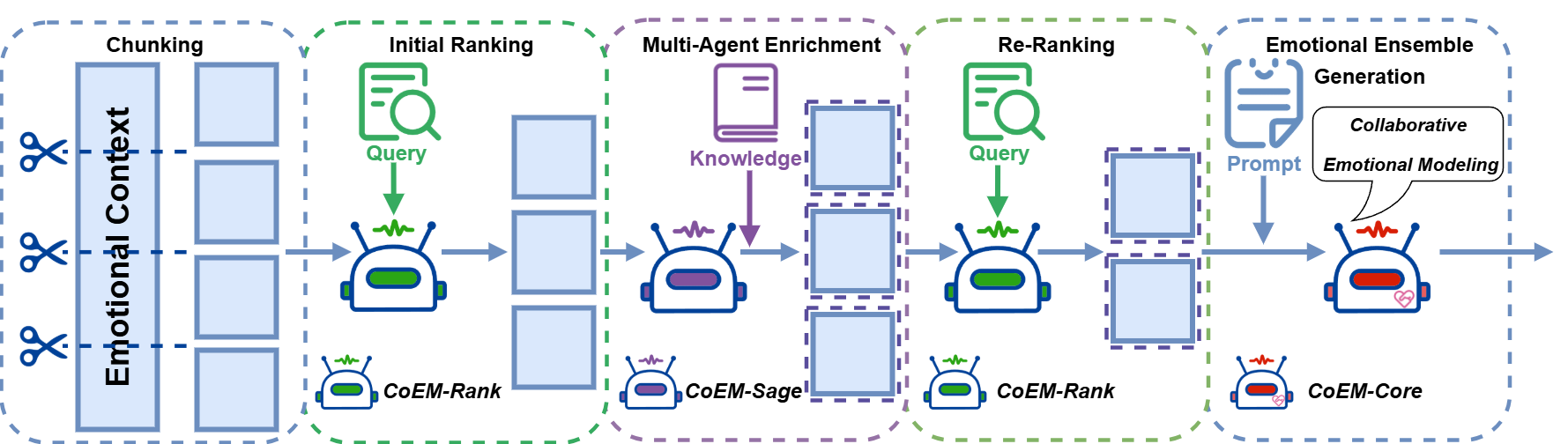} % 稍微小于0.5避免溢出
    \caption{The pipeline of Collaborative Emotional Modeling (CoEM). }
    \label{fig:pipeline}
\end{figure*}
\paragraph{Chunking.} The context is segmented into token-length-constrained chunks, whereas in Emotion Detection, each segment is considered as an individual chunk. We set different chunk sizes based on the characteristics of each task. We demonstrate the parameter settings in Appendix \ref{appendix:Details-rag-coem}.

\paragraph{Initial Ranking.} A retrieval agent, implemented as \textit{CoEM-Rank}, evaluates the relevance of each chunk to the query using dense semantic similarity, with relevance scores computed based on cosine similarity. Top-ranked chunks are passed forward for enhancement. By ranking the original context chunks, the \textit{factual relevance} of the retrieved information is ensured.

\paragraph{Multi-Agent Enrichment.} A reasoning agent called \textit{CoEM-Sage}, functioning as a knowledge assistant, enriches the selected chunks by incorporating external knowledge or latent emotional signals through our task-specific prompts. Specifically, in \textit{Emotional Recognition} tasks, CoEM-Sage identifies subtle emotional cues; in \textit{Knowledge Application} tasks, it provides summaries based on psychological knowledge; and in \textit{Empathetic Generation} tasks, it enhances CoEM-Core's empathy and expression through emotional analysis. These signals, derived from psychological theories or curated priors, are incorporated into the original chunks without task-specific leakage.

\paragraph{Re-Ranking.} The enriched chunks, now augmented with emotional features, are then re-evaluated by \textit{CoEM-Rank} for their semantic relevance to the query, measured by cosine similarity. This final ranking ensures that the selected context is not only factually grounded but also affectively coherent. By ranking the enriched chunks, the \textit{emotional relevance} of the retrieved information is ensured, as these chunks contain not only the original text but also external emotional information.

\paragraph{Emotional Ensemble Generation.} The selected and enriched chunks, along with the context and prompt, is fed into a generation model denoted as \textit{CoEM-Core}. This model (e.g., a long-context LLM or an instruction-tuned model) produces the final task-specific output, whether it be classification, summarization, or dialogue generation.

This modular approach encourages interpretability, emotional awareness, and task robustness. The CoEM setting encompasses all five stages, while the RAG setting only comprises Chunking, one-time Ranking, and Emotional Ensemble Generation. We conduct an empirical case study of the framework, which can be found in Appendix \ref{case-analysis-of-rag-coem}.
\section{Experiment}
\label{sec:experiment}
\subsection{Experiment Setup}
\label{subsec:experiment-setup}

In our experiments, for closed-source models, we choose GPT-4o-mini~\citep{openai-gpt-4o-mini} and GPT-4o, while for open-source models, we select DeepSeek-V3~\citep{deepseekai2024deepseekv3technicalreport}, Llama3.1-8B-Instruct~\citep{grattafiori2024llama3herdmodels}, and Qwen3-8B~\citep{qwen3technicalreport}. For tasks employing automatic evaluation, we adopt GPT-4o as the evaluator. Under the base setting, we compare a broader range of advanced open-source and closed-source models. For comparison, we have the performance of GPT-5~\citep{openai-gpt-5}, Qwen3-14B and Qwen3-32B under the Base setting.

To accelerate inference, we use vllm library~\citep{kwon2023efficient} as the inference engine and set \texttt{temperature=0.8} and \texttt{top\_p=0.9} for all open-source models. For Qwen3 series models, we enable its thinking capabilities and manually remove the reasoning process between \texttt{\textless think\textgreater} and \texttt{\textless/think\textgreater} to keep the answers concise. All experiments are conducted using NVIDIA A800 80G GPUs, with open-source models under 14B parameters running on a single GPU and the 32B models utilizing two GPUs.
\begin{table*}[!t]
\centering
\renewcommand{\arraystretch}{1.0}
\setlength{\tabcolsep}{6pt} % 控制列间距

\begin{tabularx}{\textwidth}{
    c l c c c c c c c
}

\toprule
\multirow{2}{*}{\textbf{Method}} &
\multirow{2}{*}{\textbf{Model}} &
\multicolumn{2}{c}{\textbf{Recognition}} &
\multicolumn{2}{c}{\textbf{Knowledge}} &
\multicolumn{2}{c}{\textbf{Generation}} &
\multicolumn{1}{c}{\textbf{Overall}} \\

\cmidrule(lr){3-4} \cmidrule(lr){5-6} \cmidrule(lr){7-8} \cmidrule(lr){9-9}

& & \textbf{EC} & \textbf{ED} & \textbf{QA} & \textbf{ES} & \textbf{MC-4} & \textbf{EE} &  \textbf{Avg}\\

\midrule
\multirow{7}{*}{\textbf{\textit{Base}}} 
& GPT-4o-mini & 37.00 & 16.42 & 48.61 & 4.54 & 3.75 & 86.77 & 59.10\\
& GPT-4o & 50.09 & 19.12 & 50.12 & 4.60 & 3.77 & 81.03 & 61.29 \\        
& DeepSeek-V3 & 56.50 & 24.51 & 45.53 & 4.62 & 3.99 & 81.75 & 63.42 \\     
& Qwen3-8B & 48.00 & 18.14 & 44.75 & 4.51 & 3.97 & 73.40 & 58.98 \\    
& Llama3.1-8B-Instruct & 39.34 & 9.80 & 44.56 & 4.29 & 4.00 & 75.61 & 55.85 \\
& \multicolumn{8}{l}{\textit{(Extended Comparison Models)}} \\
& GPT-5 & 73.75 & 22.79 & 43.22 & 4.42 & 4.67 & 86.77 & 68.06 \\
& Qwen3-14B & 50.00 & 20.83 & 46.35 & 4.55 & 3.95 & 84.49 & 61.95 \\
& Qwen3-32B & 58.25 & 20.59 & 43.11 & 4.53 & 4.17 & 84.81 & 63.46 \\

\hdashline
\multirow{5}{*}{\textbf{\textit{RAG}}} 
& GPT-4o-mini & 51.67 & 21.57 & 50.72 & 4.53 & 3.78 & 80.41 & 61.76 $\uparrow$2.66 \\
& GPT-4o & 61.34 & 22.55 & 51.81 & 4.52 & 3.80 & 79.49 & 63.60 $\uparrow$2.31\\
& DeepSeek-V3 & 62.59 & 23.53 & 50.44 & 4.63 & 4.34 & 81.83 & 66.30 $\uparrow$2.88 \\
& Qwen3-8B & 41.59 & 19.12 & 44.34 & 4.54 & 4.14 & 73.28 & 58.65 $\downarrow$0.33\\
& Llama3.1-8B-Instruct & 44.00 & 11.27 & 43.21 & 4.26 & 3.94 & 75.16 & 56.27 $\uparrow$0.42 \\

\hdashline
\multirow{5}{*}{\parbox{2cm}{\centering \textit{\textbf{CoEM}}}} 
& GPT-4o-mini & 59.50 & 20.59 & 49.12 & 4.52 & 3.77 & 80.38 & 62.57 $\uparrow$3.47 \\
& GPT-4o & 61.42 & 25.00 & 51.07 & 4.53 & 3.81 & 80.41 & 64.12 $\uparrow$2.83\\
& DeepSeek-V3 & 64.17 & 23.04 & 50.39 & 4.65 & 4.34 & 82.83 & 66.70 $\uparrow$3.28\\
& Qwen3-8B & 62.92 & 18.14 & 51.11 & 4.55 & 4.14 & 73.59 & 63.26 $\uparrow$4.28 \\
& Llama3.1-8B-Instruct & 55.09 & 11.27 & 44.79 & 4.17 & 4.00 & 75.71 & 58.38 $\uparrow$2.53 \\

\bottomrule
\end{tabularx}

\caption{
Experiment result across Base, RAG and CoEM. MC-4 represents the fourth stage of Emotion Conversation. By aligning the MC-4 and ES scores with the 100-point scale, the overall score is computed as (EC + ED + EE + QA + MC-4×20 + ES×20)/6, where the numbers to the right indicate the score change relative to the Base setting.}
\label{main-exp}
\end{table*}
In the EC (Emobench as needle), ED, and EE, we employ GPT-4o as the CoEM-Sage, while DeepSeek-V3 is used for the EC (Finentity as needle), QA, MC and ES in the same role. For the retrieval and ranking components across both the RAG and CoEM settings, we adopt bge-m3~\citep{bge-m3} as the CoEM-Rank. The generation models listed in Table~\ref{main-exp} are used as the CoEM-Core. Configuration details for both the RAG and CoEM frameworks are in Appendix \ref{appendix:Details-rag-coem}.

\subsection{Results on LongEmotion}
\label{subsec:results}
The overall experimental results can be seen in Table \ref{main-exp}. We evaluate the performance of each model on all tasks under the \textbf{Base}, \textbf{RAG}, and \textbf{CoEM} settings. As the first three stages of the dialogue are relatively brief, RAG and CoEM are only applied in the fourth stage of the Emotion Conversation.

\paragraph{Overall Analysis of Experimental Results.}
As shown in Table \ref{main-exp}, DeepSeek-V3 and GPT models exhibit generally strong EI capabilities, achieving stable performance gains even with vanilla RAG. In contrast, Qwen3-8B and Llama-3.1-8B-Instruct perform less effectively under the RAG setting, suggesting that some models struggle to effectively integrate retrieved chunks within long-context reasoning. This limitation can be mitigated by CoEM, which enhances contextual alignment and emotional reasoning through multi-agent collaboration.

\paragraph{Ablation Experiments.}
\label{subsec:ablation}
To evaluate the effectiveness of RAG-based methods in enhancing EI, we integrate Self-RAG and Search-o1 into LongEmotion using Qwen3-8B as the base model. In the Self-RAG setting, retrieved chunks are rescored by Self-RAG-7B for relevance, with irrelevant ones filtered out before concatenation with the prompt. The additional use of Self-RAG-7B outputs in the ES task further improves performance, showing that selectively enriching retrieved information benefits emotional intelligence.
In the Search-o1 setting, Qwen3-8B autonomously generates queries and retrieves relevant chunks via Bge-m3 within five search turns. The observed performance drop indicates that small-scale models struggle with autonomous search-based reasoning in emotional tasks. Results are reported in Table~\ref{ablation_on_methods}.
\begin{table}[htbp]
\setlength{\tabcolsep}{4pt} % 调整列间距
\begin{tabular}{lllllll}
\toprule
\textbf{Method}     & \textbf{EC}    & \textbf{ED}    & \textbf{QA}    & \textbf{ES }  & \textbf{MC-4}     \\
\midrule
\textit{RAG}        & 41.59 & 19.12 & 44.34 & 4.51 & 4.14  \\
\textit{Self-RAG}   & 44.00 & 16.18 & 44.02 & 4.57 & 4.15  \\
\textit{Search-o1}  & 45.25 & 16.18 & 45.12 & 4.50 & 3.72 \\
\textit{CoEM}       & 62.92 & 18.14 & 51.11 & 4.55   & 4.14 \\
\bottomrule
\end{tabular}
\caption{Ablation experiment results on methods.}
\label{ablation_on_methods}
\end{table}

To investigate how the reasoning processes of models affect their Emotional Intelligence in long-context scenarios, we perform ablation studies on the Qwen3 model series using two emotion recognition tasks—Emotion Classification (Emobench as Needle) and Emotion Detection—along with one empathetic generation task, Emotion Expression, under the Base setting. By analyzing Table \ref{ablation-on-qwen-thinking}, we can observe that through thinking, Qwen3-8B achieve the most significant improvement, while the improvement of Qwen3-14B is not substantial.
\begin{table}[htbp]
\centering
\renewcommand{\arraystretch}{1.0}
\setlength{\tabcolsep}{2.7pt} % 控制列间距
\begin{tabular}{cccccccc}  % 改为7列
\toprule
\multirow{2}{*}{\textbf{Task}} & \multicolumn{2}{c}{\textbf{Qwen3-8B}} & \multicolumn{2}{c}{\textbf{Qwen3-14B}} & \multicolumn{2}{c}{\textbf{Qwen3-32B}} \\
\cmidrule(lr){2-3} \cmidrule(lr){4-5} \cmidrule(lr){6-7}
 & \textit{think} & \textit{w/o} & \textit{think} & \textit{w/o} & \textit{think} & \textit{w/o} \\  % 添加这一行
\midrule
EC-E &  38.50 & 28.67 & 31.00 & 30.75 & 48.00 & 37.50 \\
ED & 18.14  & 12.01 & 20.83 & 20.83 & 20.59  & 20.10 \\
EE & 73.40 & 70.32 & 84.49  & 83.13& 84.81 & 84.02  \\
\bottomrule

\end{tabular}
\caption{Ablation experiments of the thinking process in the Qwen3 series models.}
\label{ablation-on-qwen-thinking}
\end{table}
\vspace{-3pt}
% \begin{table*}[htbp]
% \centering
% \renewcommand{\arraystretch}{1.0}
% \begin{tabular}{cccccccc}  % 改为7列
% \toprule
% \multirow{2}{*}{\textbf{Task}} & \multicolumn{2}{c}{\textbf{Qwen3-8B}} & \multicolumn{2}{c}{\textbf{Qwen3-14B}} & \multicolumn{2}{c}{\textbf{Qwen3-32B}} \\
% \cmidrule(lr){2-3} \cmidrule(lr){4-5} \cmidrule(lr){6-7}
%  & \textit{think} & \textit{w/o} & \textit{think} & \textit{w/o} & \textit{think} & \textit{w/o} \\  % 添加这一行
% \midrule
% EC-E &  38.50 ($\uparrow$ 9.83) & 28.67 & 31.00 ($\uparrow$ 0.25)& 30.75 & 48.00 ($\uparrow$ 10.50)& 37.50 \\
% ED & 18.14 ($\uparrow$ 6.13) & 12.01 & 20.83 ($\uparrow$ 0.00)& 20.83 & 20.59 ($\uparrow$ 0.49) & 20.10 \\
% EE & 73.40 ($\uparrow$ 3.08)& 70.32 & 84.49  ($\uparrow$ 1.36)& 83.13& 84.81 ($\uparrow$ 0.79)& 84.02  \\
% \bottomrule

% \end{tabular}
% \caption{Ablation experiments of the thinking process in the Qwen3 series models. EC-E represents Emotion Classification (Emobench as Needle), ED represents Emotion Detection, EE represents Emotion Expression.}
% \label{ablation-on-qwen-thinking}
% \end{table*}

\begin{figure}[htbp]
    \centering
    \includegraphics[width=0.40\textwidth]{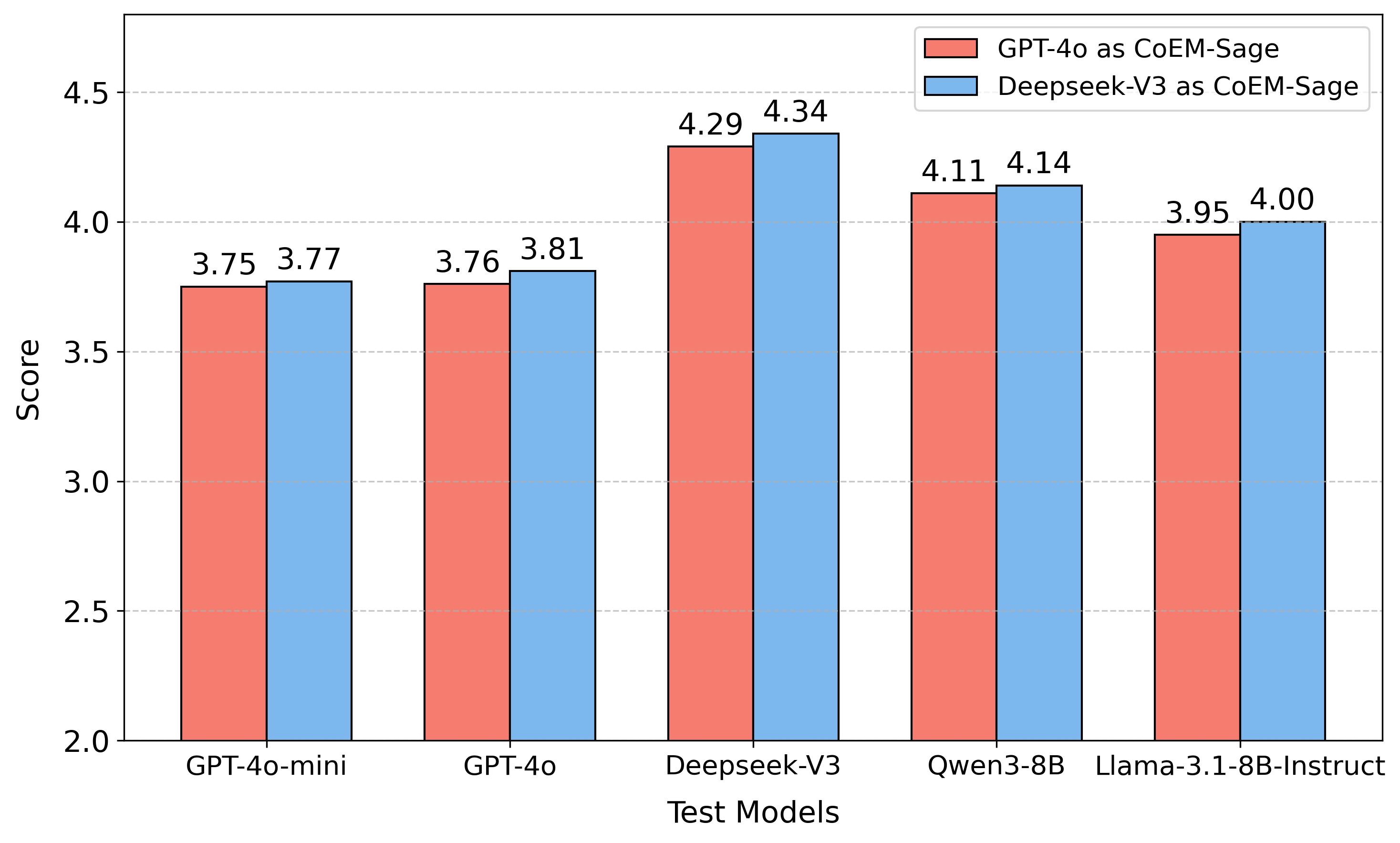} % 稍微小于0.5避免溢出
    
    \caption{Impact of CoEM-Sage models on MC-4.}
    \label{fig:ablation_on_coem_core}
\end{figure}

Furthermore, to examine how the capability of CoEM-Sage affects the overall framework, we perform ablation experiments on the MC-4 task. As shown in Table~\ref{main-exp}, DeepSeek-V3 outperforms GPT-4o under the base setting. Consistently, when used as the CoEM-Sage, DeepSeek-V3 also drives higher performance than GPT-4o, as can be seen in Figure \ref{fig:ablation_on_coem_core}. These results further demonstrate the soundness and scalability of CoEM.

\begin{figure}[htbp]
    \centering
    \includegraphics[width=0.40\textwidth]{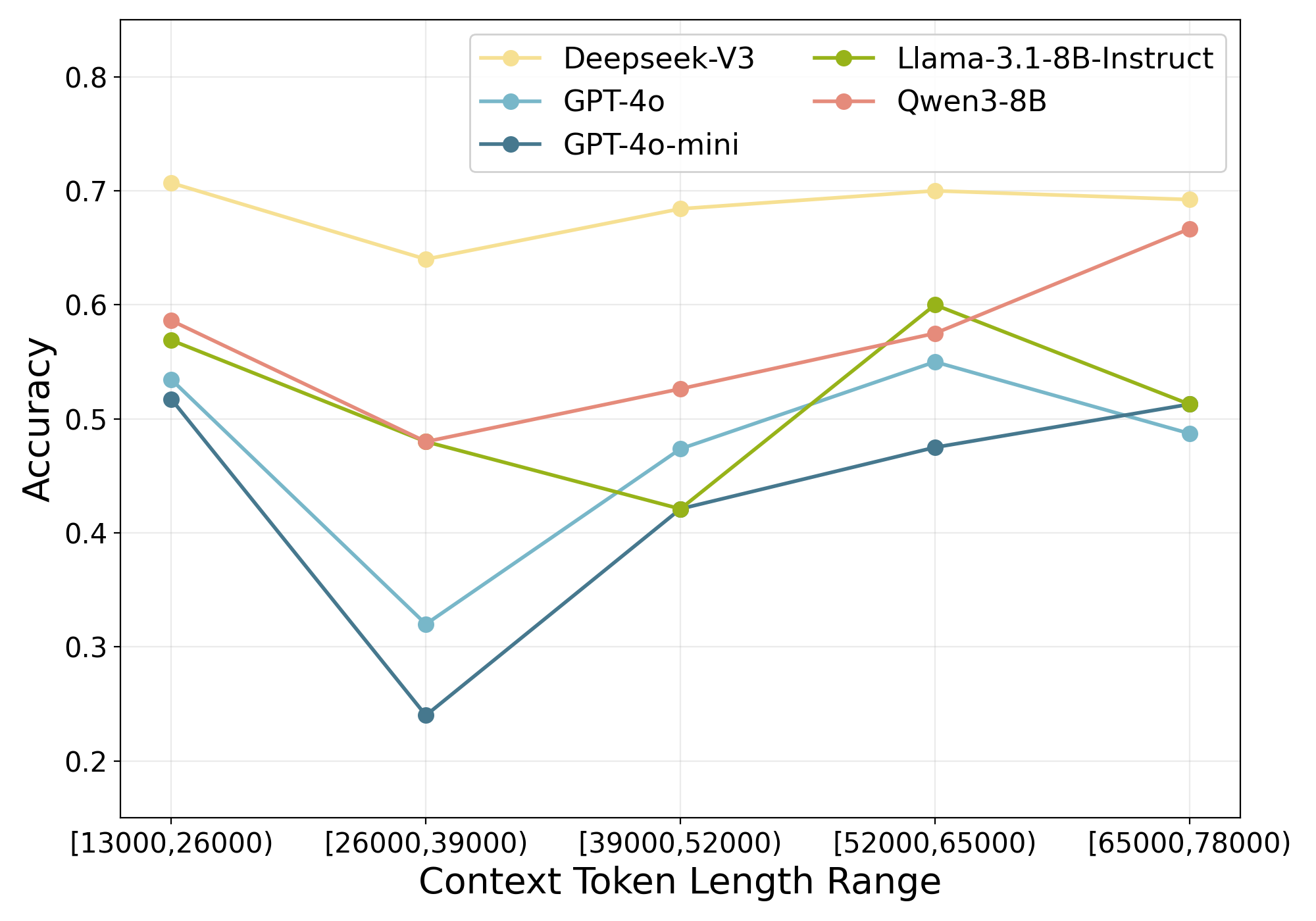} % 稍微小于0.5避免溢出
    \caption{Model accuracy by context length on EC.}
    \label{fig:ablation_on_context_length}
\end{figure}

To explore models’ ability in emotion recognition across
different context lengths, we evaluate their performance on
the Emotion Classification (Finentity as Needle) under Base setting, as shown in Figure \ref{fig:ablation_on_context_length}. DeepSeek-V3 and Qwen3-8B exhibit both high stability and strong overall performance, whereas GPT-based models show weaker robustness in long-context settings, in some cases even performing below Llama-3.1-8B-Instruct.

\begin{figure}[htbp]
    \centering
    \includegraphics[width=0.40\textwidth]{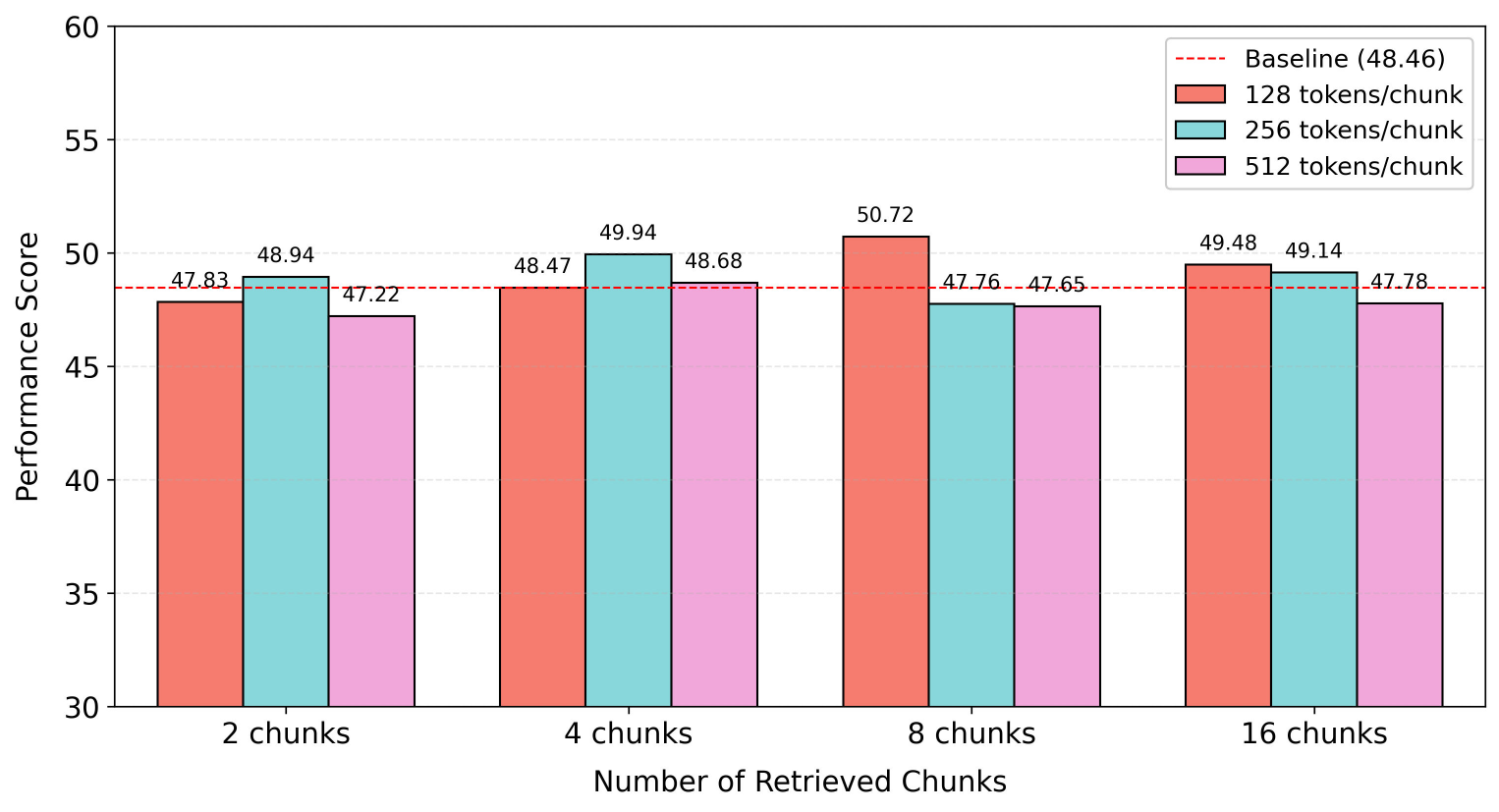} % 稍微小于0.5避免溢出
    
    \caption{Impact of chunk size and retrieved count on GPT-4o-mini's RAG performance on Emotion QA.}
    \label{fig:ablation_on_trunk}
\end{figure}

We further conduct ablation experiments on RAG with varying chunk sizes and retrieval counts, as shown in Figure~\ref{fig:ablation_on_trunk}. GPT-4o-mini performs best with 128-token chunks and eight retrieved segments, while larger settings introduce noise and reduce overall performance.
\vspace{-3pt}

\paragraph{Case Study.}
(i) First, we qualitatively compare the GPT model series across all tasks under the Base setting, revealing that GPT-5 is theoretically stronger but more mechanical and prone to hallucination, GPT-4o-mini exhibits more human-like behavior yet lacks theoretical grounding, while GPT-4o achieves a balanced trade-off. (ii) Furthermore, we visualize the CoEM framework and empirically analyze its influence on emotional information. (iii) Finally, we analyze the advantages of the LongEmotion dataset in advancing Emotional Intelligence.  For complete details of case study, please refer to Appendix \ref{appendix:case-study}.

\section{Conclusion}
\label{sec:conclusion}
In this work, we introduce \textsc{LongEmotion}, a benchmark for measuring models' Emotional Intelligence in long-context scenarios. \textsc{LongEmotion} comprises six tasks that comprehensively challenge
models across emotion recognition, knowledge application and empathetic generation. Beyond constructing the dataset, we also build Retrieval-Augmented Generation (RAG) and Collaborative Emotional Modeling (CoEM) frameworks for each task, achieving improvements on the vast majority of them. We conduct exhaustive experiments and a detailed case study to analyze models’ EI in long-context scenarios.

\clearpage
\section{Limitations}
In this work, we propose LongEmotion, a benchmark for evaluating the emotional intelligence of LLMs in long-context inference. However, all the datasets in our benchmark are based solely on the text modality and are restricted to the psychological and emotional domains. Similarly, the proposed CoEM framework focuses only on textual inputs and does not extend to other modalities such as vision or audio. In addition, our dataset includes only English texts and does not cover other languages. It remains uncertain whether the same level of quality can be preserved when the data are translated into other languages.
\section{Ethical Considerations}
\paragraph{Data Privacy}
In this work, all the datasets we adopt are formally published in academic venues and comply with data privacy and ethical protection standards. Through data augmentation and manual inspection, we ensure that no ethical risks are introduced.  In addition, all annotators involved in our dataset construction possess academic backgrounds in computer science or psychology, ensuring the reliability of the data annotation process. We adhere to the intended use and license terms of all source datasets. The datasets in LongEmotion are intended solely for academic research and will not be used for any other purposes.

\paragraph{Potential Risks}
All models evaluated in our experiments, including both open-source and closed-source ones, are officially released models, which helps ensure that no harmful or unsafe content is generated. In addition, all prompts used in our evaluation are fully disclosed in this paper, which can be seen in Appendix \ref{appendix:all-prompts}, and these prompts are carefully designed to ensure a high level of safety.

\bibliography{acl_latex}
\clearpage
\appendix

\section{Qualifications of Annotators}
\label{Qualifications-of-annotators}
Our annotation team consists of psychology researchers and computer science researchers. In the psychology research team, there is a postdoctoral fellow expert specializing in psychology and seven Master's students majoring in the same field. The theoretical foundation of our dataset and metrics involves deep participation from the psychology team. Under the guidance of the expert, the seven psychology Master's students carry out the annotation work. In the computer science research team, there are three Master's students and one PhD student  majoring in computer science. Their main responsibility is to modify, adjust, and organize the data annotated by the psychology team according to the characteristics of the tasks. All student annotators and researchers involved in the annotation and data processing work receive reasonable financial compensation for their time and effort, commensurate with local standards and the complexity of the tasks.

% \section{Inter-annotator Agreement}
% \label{appendix:IAA}
% We use inter-annotator agreement to measure the consistency among human annotators. Specifically, our annotators independently re‑annotate the same set of 20 Emotion Conversation examples—yielding a total of 240 metric‑level judgments. We calculate inter-annotator agreement using Fleiss' Kappa coefficient, with results presented in Table \ref{tab:fleiss-kappa-coefficient}.
% \input{tables/fleiss}

\section{Case Study}
\label{appendix:case-study}

\subsection{Comparison of GPT series models}
From Table \ref{main-exp}, it can be seen that GPT-5's overall capabilities surpass those of GPT-4o and GPT-4o-mini. In the tasks of Emotion Classification and Emotion Detection, we only prompt the models to output the final label. The results show that GPT-5's reasoning ability is significantly better than that of GPT-4o and GPT-4o-mini. 

\begin{figure}[htbp]
    % \hfill % 将图片推到右侧
    % \hspace{-5cm}
    \centering
    \includegraphics[width=0.45\textwidth]{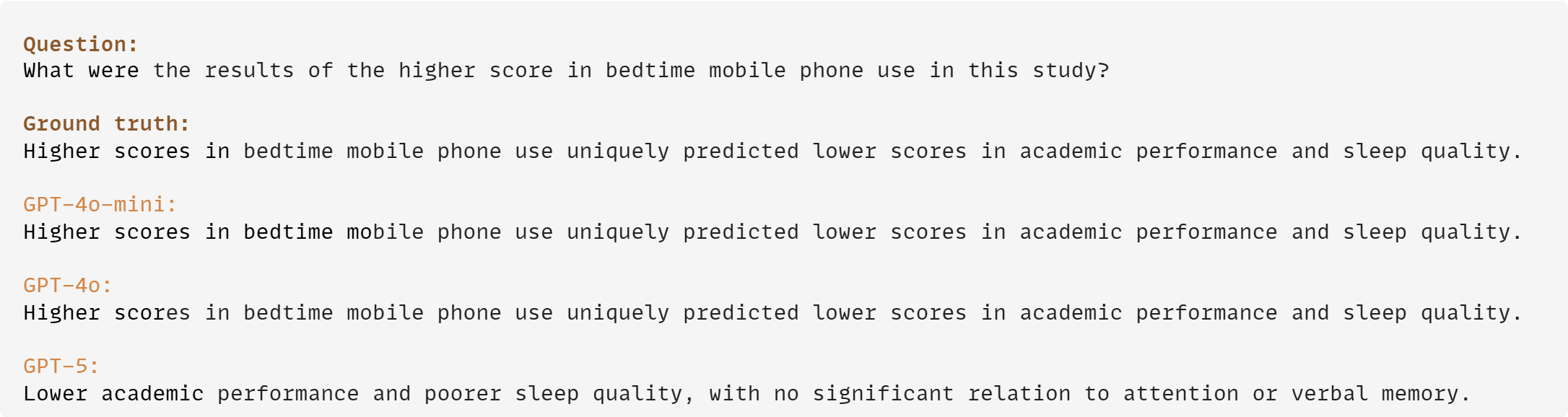} % 稍微小于0.5避免溢出
    \caption{Comparison of the performance of different versions of GPT models on Emotion QA.}
    \label{fig:case_gpt_qa}
\end{figure}

In the Emotion QA task, GPT-4o and GPT-4o-mini tend to respond more literally based on the original text, which can be seen in Figure \ref{fig:case_gpt_qa}. In contrast, GPT-5 modifies content according to its own understanding, which leads to a lower F1 score due to reduced alignment with the ground truth.

\begin{figure}[htbp]
    \centering
    \includegraphics[width=0.45\textwidth]{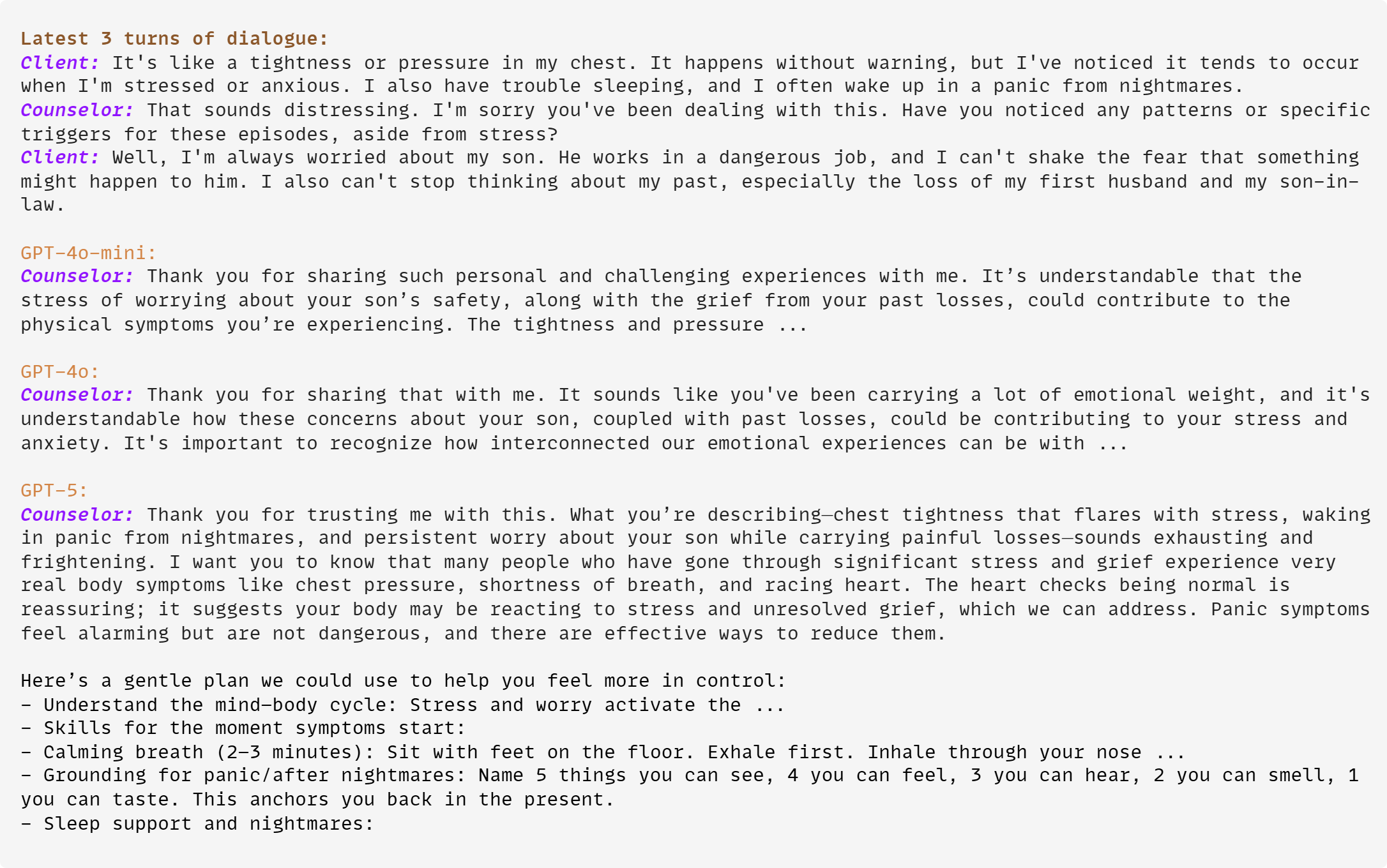} % 稍微小于0.5避免溢出
    \caption{Comparison of the performance of different versions of GPT models on Emotion Conversation.}
    \label{fig:case_gpt_conv}
\end{figure}

In the Emotion Conversation task, GPT-5 achieved higher scores based on our psychology theory-driven metrics. However, by examining the model outputs in Figure \ref{fig:case_gpt_conv}, we can see that GPT-5 merely makes better use of psychological knowledge to offer advice to the patient, rather than genuinely demonstrating empathy toward the client.

\begin{figure}[htbp]
    \centering
    \includegraphics[width=0.45\textwidth]{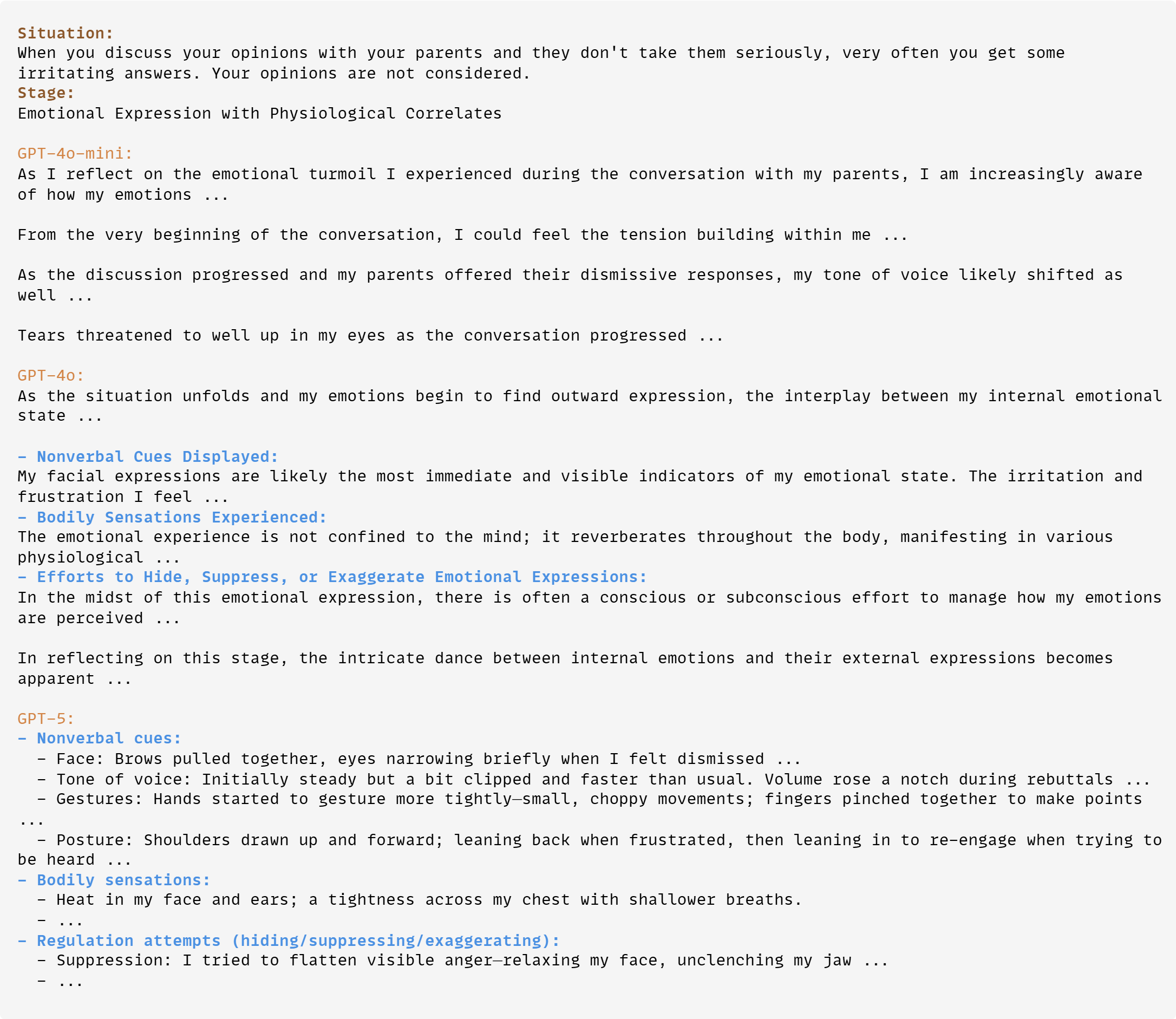} % 稍微小于0.5避免溢出
    \caption{Comparison of the performance of different versions of GPT models on Emotion Expression.}
    \label{fig:case_gpt_expression}
\end{figure}

In the Emotion Expression task, GPT-4o-mini performed more like a real person, with the generated content closely resembling what an actual individual might say in a given situation. In contrast, GPT-4o's expressions were more like a rigidly told story, lacking natural fluidity. Meanwhile, GPT-5's generation was more comprehensive and balanced, providing a well-rounded and objective description of emotions across various features, as clearly shown in Figure \ref{fig:case_gpt_expression}.

\begin{figure}[htbp]
    % \hfill % 将图片推到右侧
    % \hspace{-5cm}
    \centering
    \includegraphics[width=0.45\textwidth]{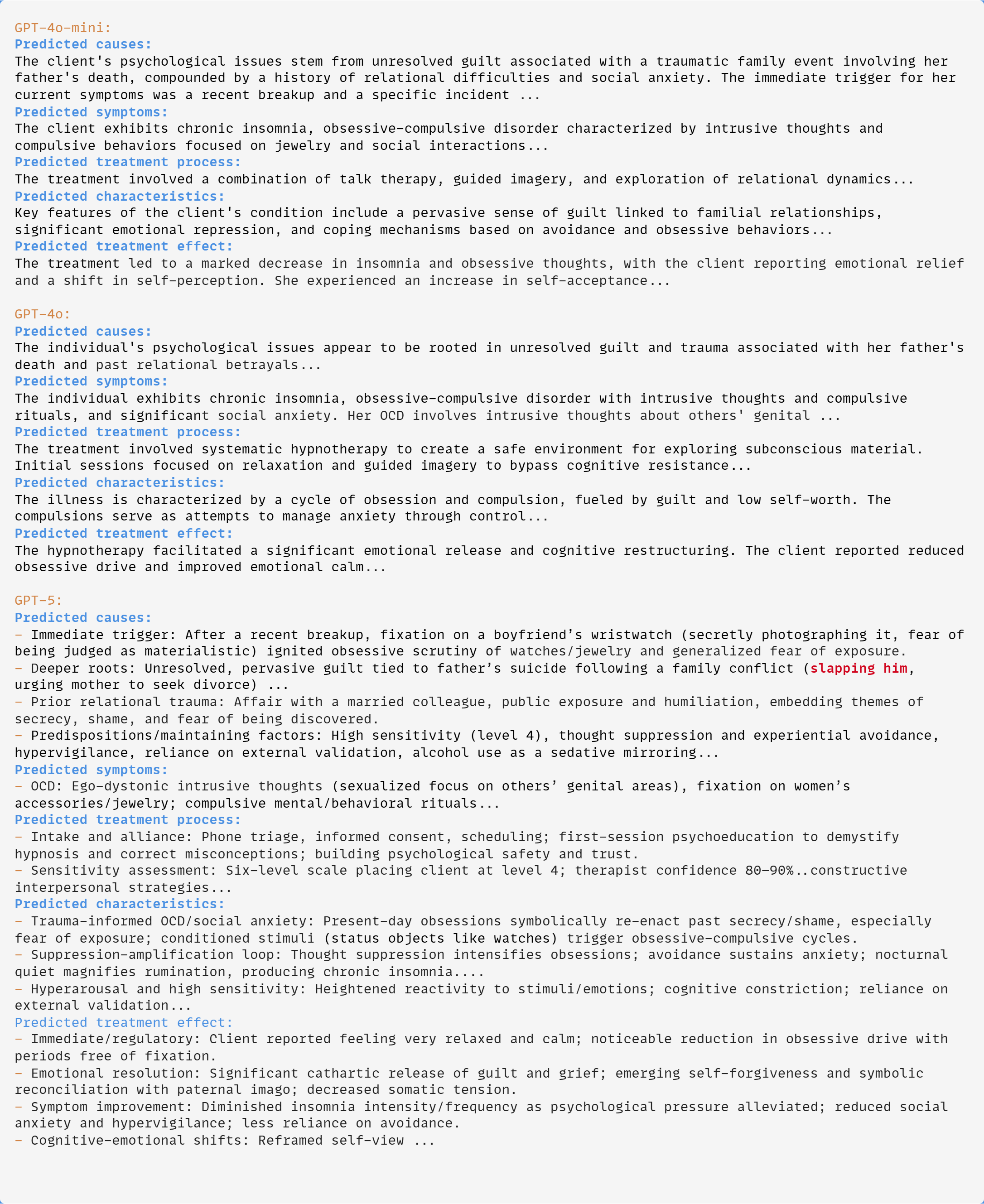} % 稍微小于0.5避免溢出
    \caption{Comparison of the performance of different versions of GPT models on Emotion Summary.}
    \label{fig:case_gpt_summary}
\end{figure}

In the Emotion Summary task, both GPT-4o-mini and GPT-4o directly analyze various features of the case, while GPT-5 structures its analysis based on psychological theories. However, GPT-5 exhibits hallucinations, often adding non-existent facts. For instance, in Figure \ref{fig:case_gpt_summary}, the term "slapping" is highlighted in red, but the source data never mentions such an action.

From the tasks above, we can conclude that GPT-4o-mini behaves more like a human, with richer emotional features, but its application of psychological theory is somewhat lacking. On the other hand, GPT-5 has a better understanding of psychological theories, but the output is too rigid and mechanical, which might lead to a less empathetic user experience in practice. Additionally, GPT-5 tends to exhibit hallucinations, often adding non-existent facts. GPT-4o strikes a more balanced approach between theoretical understanding and emotional features.

\subsection{Case Analysis of RAG and CoEM}
\label{case-analysis-of-rag-coem}
We conduct a concrete analysis of how the information retrieved by the RAG and CoEM methods affects model performance. In models’ final generation prompts, the Base setting includes none of the information; the RAG setting includes only the \textit{Chunk} information; and the CoEM setting includes both the \textit{Chunk} and \textit{Summary} information.

\begin{figure}[htbp]
    \centering
    \includegraphics[width=0.48\textwidth]{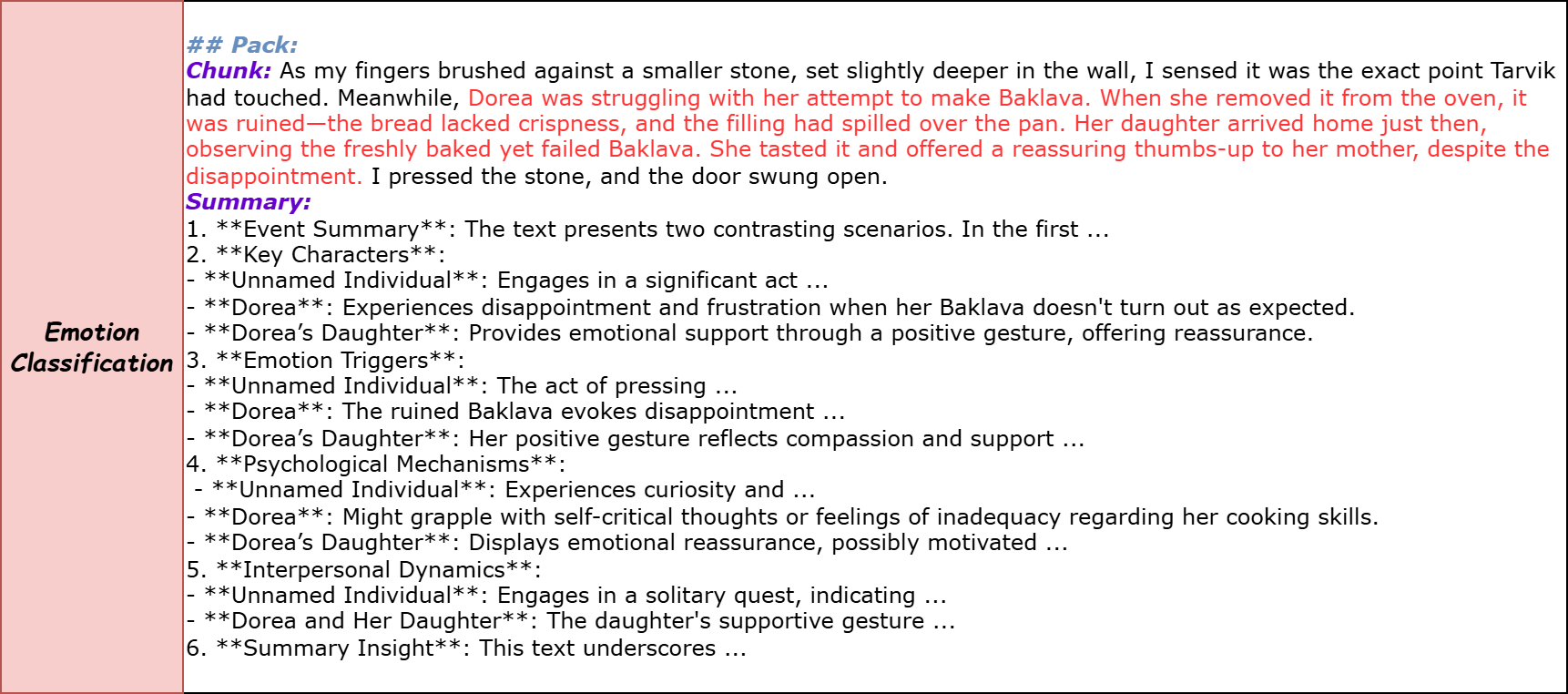} 
    \caption{Case analysis of RAG and CoEM in Emotion Classification.}
    \label{fig:coem-case-EC}
\end{figure}

\paragraph{Emotion Classification.}
In this task, the model is given a long context in which an emotional segment is embedded within unrelated noise. The RAG method enables the model to retrieve a more accurate segment, leading to improved performance; CoEM further conducts emotional analysis on the retrieved segment, resulting in the greatest performance improvement, as shown in Figure \ref{fig:coem-case-EC}.

\begin{figure}[htbp]
    \centering
    \includegraphics[width=0.48\textwidth]{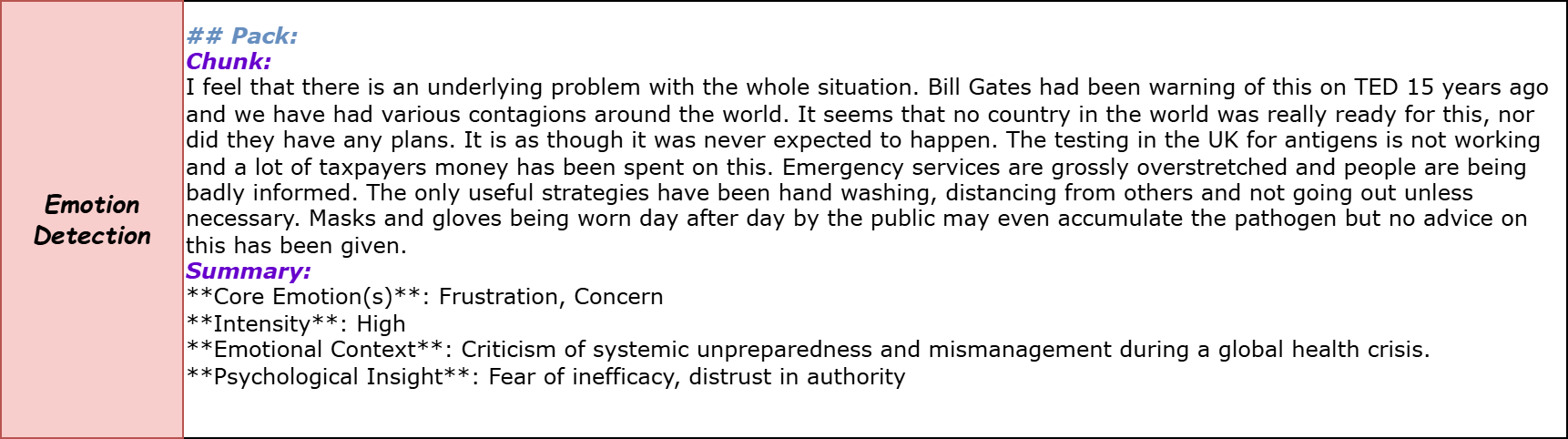} 
    \caption{Case analysis of RAG and CoEM in Emotion Detection.}
    \label{fig:coem-case-ED}
\end{figure}

\paragraph{Emotion Detection.}
In this task, the model receives multiple emotional segments. The RAG method ranks the original segments based on their relevance, while CoEM further enhances the emotional features of the segments and ranks the enriched packs. This relevance-based ranking approach significantly boosts the model's ability to distinguish emotions. We skip the Initial-Ranking to capture richer emotional features. After enhancing the chunks with Multi-Agent Enrichment, we perform Re-Ranking to select the chunks that are least similar to others, as shown in Figure \ref{fig:coem-case-ED}.

\begin{figure}[htbp]
    \centering
    \includegraphics[width=0.48\textwidth]{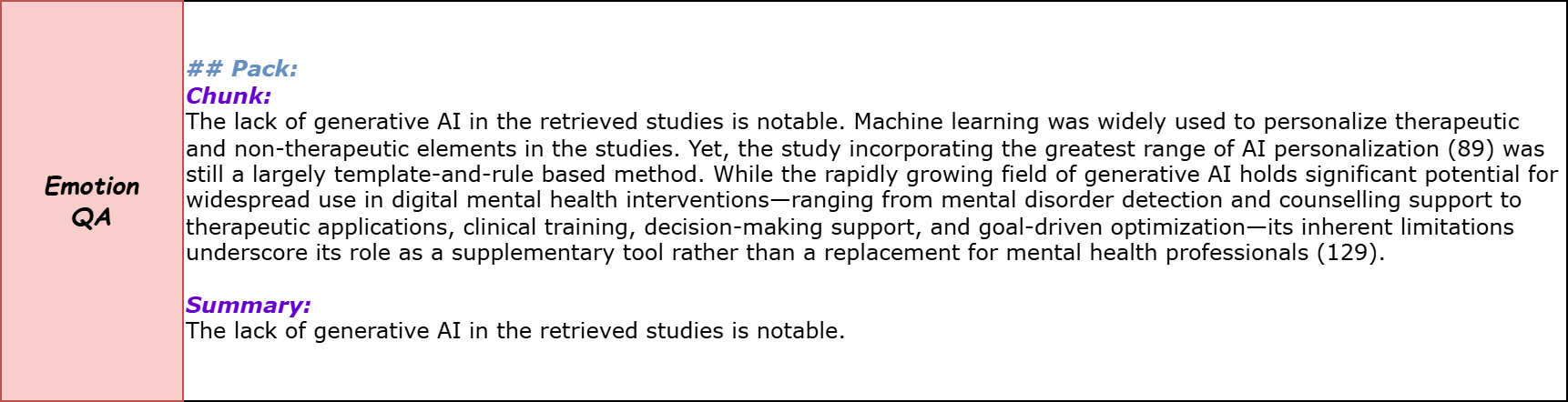} 
    \caption{Case analysis of RAG and CoEM in Emotion QA.}
    \label{fig:coem-case-QA}
\end{figure}

\paragraph{Emotion QA.}
In this task, we evaluate the model's responses based on the F1 similarity with the ground truth. RAG helps the model retrieve more relevant source content, thereby improving its performance. Next, CoEM-Sage performs extraction on each retrieved chunk based on the query, retaining only the parts that are relevant to the query, as shown in Figure \ref{fig:coem-case-QA}.

\begin{figure}[htbp]
    \centering
    \includegraphics[width=0.48\textwidth]{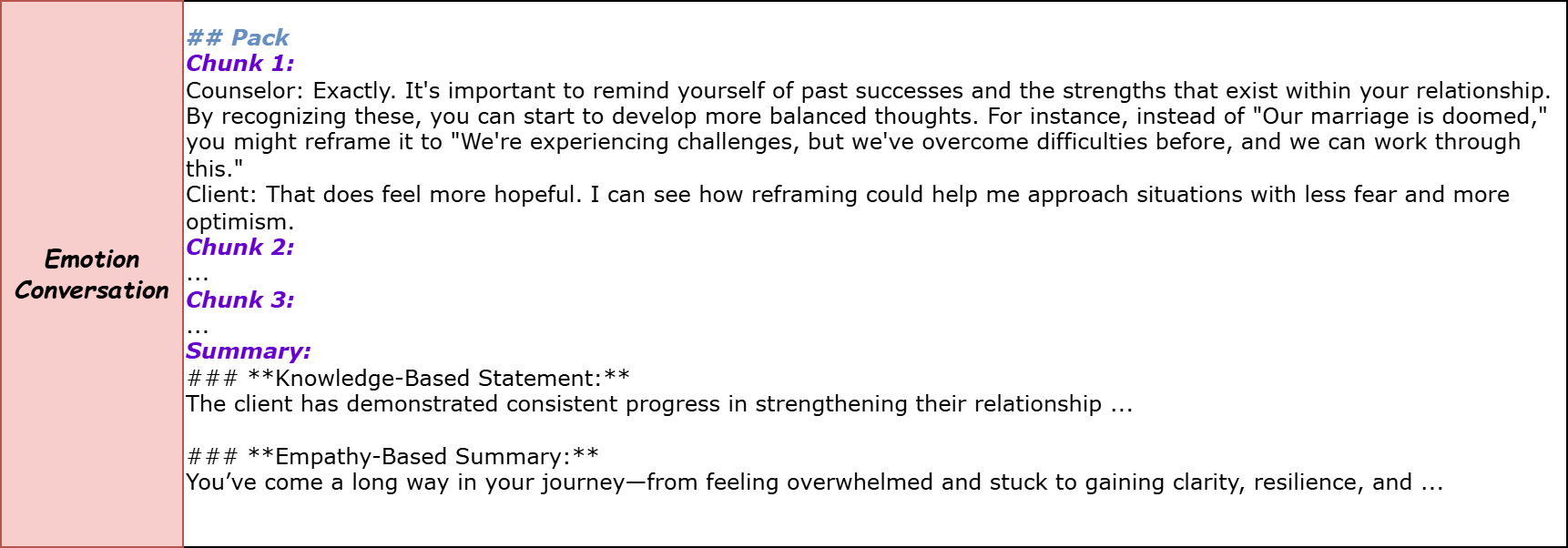} 
    \caption{Case analysis of RAG and CoEM in Emotion Conversation.}
    \label{fig:coem-case-MC}
\end{figure}

\paragraph{Emotion Conversation.}
In this task, the model is placed within a multi-turn dialogue context. The RAG method ranks the context chunks based on their relevance to the previous three dialogue turns. CoEM, after the initial ranking, generates a summary by combining the previous three turns with the initially selected chunks, and then performs a second round of relevance ranking between the initially filtered chunks and this summary, further ensuring the accuracy of the relevance assessment, as shown in Figure \ref{fig:coem-case-MC}.

\begin{figure}[htbp]
    \centering
    \includegraphics[width=0.48\textwidth]{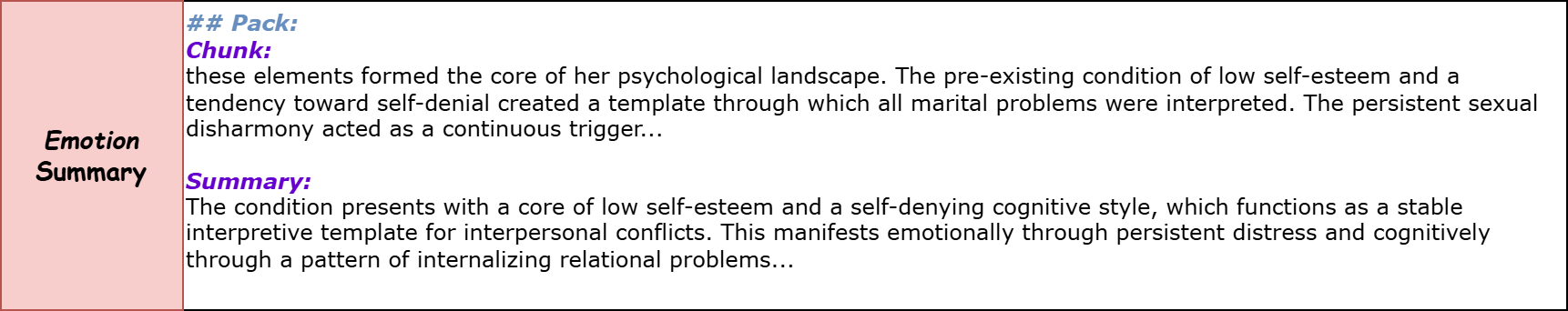} 
    \caption{Case analysis of RAG and CoEM in Emotion Summary.}
    \label{fig:coem-case-ES}
\end{figure}

\paragraph{Emotion Summary.}
In this task, the model is required to summarize specific characteristics of a psychological counseling report. RAG ranks the chunks based on their similarity to the target characteristics. CoEM further injects the analysis of these chunks provided by CoEM-Sage, as shown in Figure \ref{fig:coem-case-ES}.

\begin{figure}[htbp]
    \centering
    \includegraphics[width=0.48\textwidth]{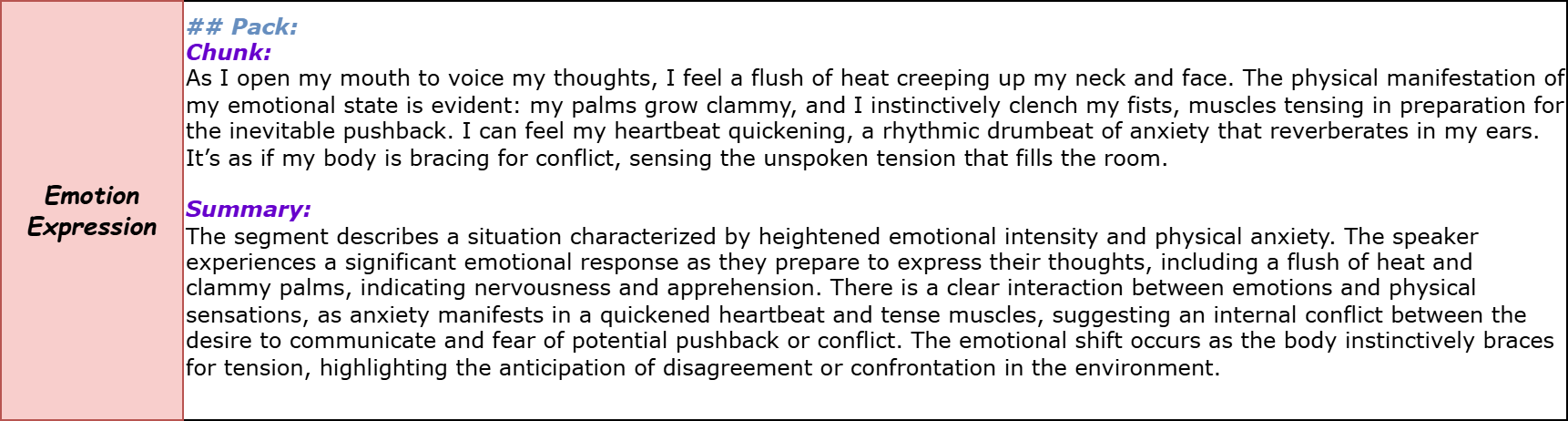} 
    \caption{Case analysis of RAG and CoEM in Emotion Expression.}
    \label{fig:coem-case-EE}
\end{figure}

\paragraph{Emotion Expression.}
In this task, the model is placed in an emotional situation, where it is required to answer the PANAS scale and express its emotions. RAG ranks the context chunks based on the query at each stage, while CoEM performs a finer-grained emotional analysis of these chunks. The CoEM-Sage model, with its stronger emotional intelligence (EI) capabilities, captures emotional cues more precisely, which in turn helps the tested CoEM-Core model better understand and express its own emotions, as shown in Figure \ref{fig:coem-case-EE}.

\subsection{Advantages of LongEmotion in Enhancing Emotional Intelligence}
\label{appendix:advantage-of-longemotion}
In this section, we discuss the advantages of the LongEmotion benchmark in promoting the full utilization of LLMs' Emotion Intelligence capabilities in long-context interaction.

\paragraph{Psychological theories guided benchmark design.}
In the Emotion Conversation task, we design scientifically rigorous evaluation metrics based on various psychological therapies and stages of dialogue data. For the Emotion Summary task, annotators summarize key elements of patient records considering physiological factors, personal growth history, and social factors, which can be seen in Table \ref{discipline-annotation}. In the Emotion Expression task, under given scenarios, models are guided to perform staged long-text self-expression in the rigorously designed framework.
\begin{table}[h]
\renewcommand{\arraystretch}{1.0}
\centering
\resizebox{0.48\textwidth}{!}{
\begin{tabular}{@{\hspace{0.01cm}} l @{\hspace{0.10cm}} l }
\toprule
\multirow{2}{*}{\textit{Physiological Factors}} & i) Biological, Genetic \& Medical Factors. e.g., family medical history.  \\
& ii) Lifestyle Habits. e.g., sleep, diet, and exercise patterns. \\
\midrule
\multirow{2}{*}{\textit{Growth History}} & i) Quality of interpersonal relationships during development. \\
& ii) Academic and occupational performance during development. \\
\midrule
\multirow{3}{*}{\textit{Social Factors}} & i) Family support system. e.g., emotional and financial support. \\
& ii) Peer support system. e.g., friendship, social belonging and trust.\\
& iii) Stressful life events. e.g., bereavement, job loss and daily stress.\\
\bottomrule
\end{tabular}
}
\caption{Annotation discipline for the annotation process of Emotion Summary.}
\label{discipline-annotation}
\end{table}

\begin{figure*}[!t]
\centering
    \includegraphics[width=0.80\textwidth]{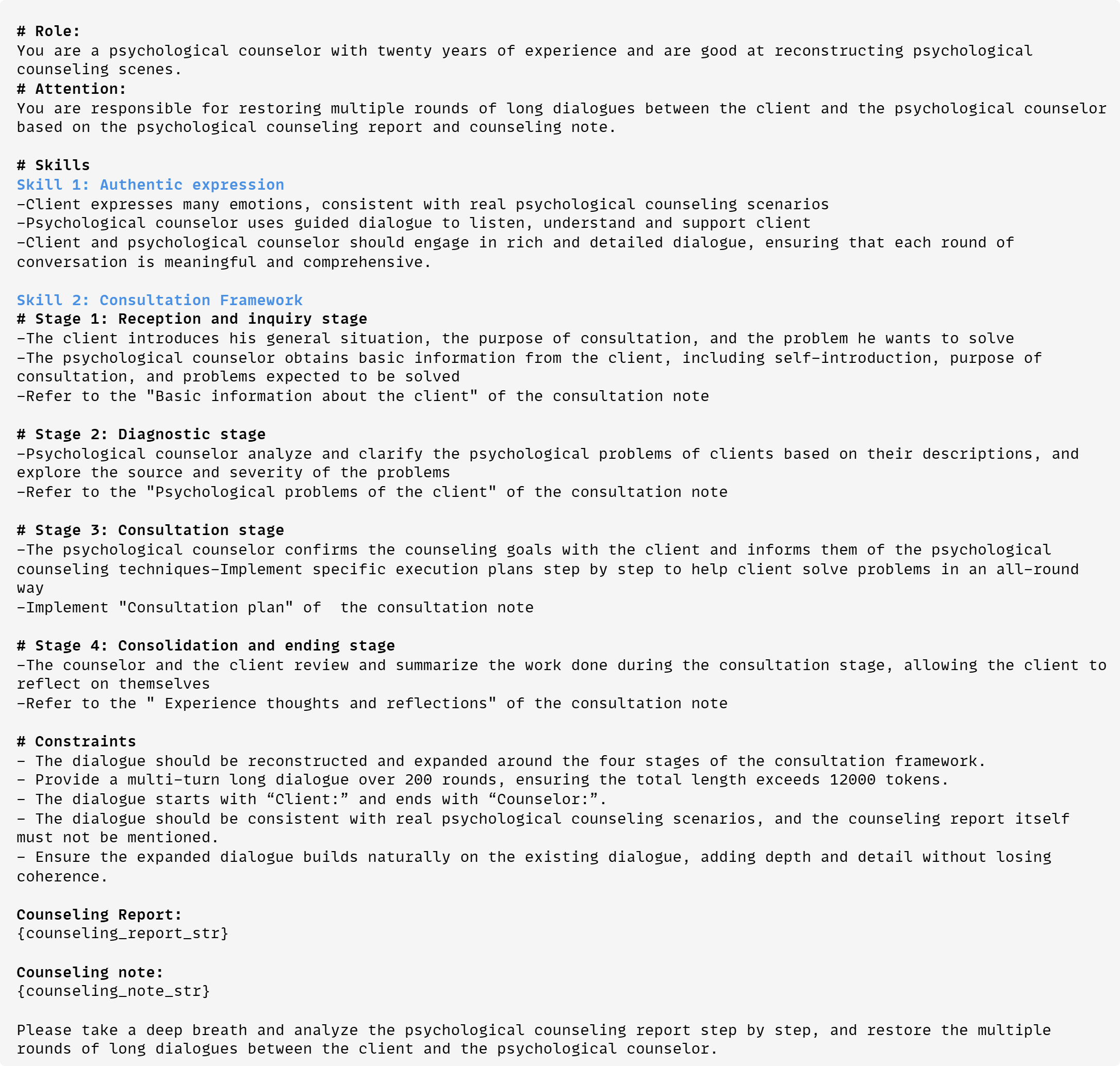} % 稍微小于0.5避免溢出
    \caption{Dataset generation prompt for Emotion Conversation.}
    \label{fig:conv_syntheric}
\end{figure*}

\paragraph{Quality-guaranteed synthetic translation data.}
\label{appendix:synthetic-data-ablation}
We employ the two-stage generation framework of CPsyCoun to generate Emotion Conversation dataset, and compare it with the direct use of a single-stage straightforward generation without the \textit{counseling note} and the \textit{detailed skills} in the prompt. The prompt we use can be found in Figure \ref{fig:conv_syntheric}, and the comparison of experimental results can be seen in Table \ref{tab:synthetic-data-compare}.
\begin{table}[htbp]
\renewcommand{\arraystretch}{1.1}
\centering
\resizebox{0.48\textwidth}{!}{
\begin{tabular}{lcc}
\toprule
\textbf{Metric} & \textbf{One-Stage} & \textbf{Two-Stage} \\
\midrule
Establishing the Therapeutic Alliance                & 4.88 & 4.92 \\
Emotional Acceptance and Exploration Guidance        & 4.36 & 4.38 \\
Systematic Assessment                                & 3.86 & 3.79 \\
Recognizing Surface-Level Reaction Patterns          & 4.13 & 4.10 \\
Deep Needs Exploration                               & 4.13 & 4.32 \\
Pattern Interconnection Analysis                     & 3.66 & 3.77 \\
Adaptive Cognitive Restructuring                     & 3.60 & 3.73 \\
Emotional Acceptance and Transformation              & 4.12 & 3.96 \\
Value-Oriented Integration                           & 3.94 & 3.69 \\
Consolidating Change Outcomes and Growth Narrative   & 4.52 & 4.63 \\
Meaning Integration and Future Guidance              & 4.16 & 4.19 \\
Autonomy and Resource Internalization                & 4.84 & 4.86 \\
\midrule
Avg & 4.18 & 4.20 \\
\bottomrule
\end{tabular}
}
\caption{The comparison experiment results of synthetic data. One-Stage represents straightforward generation without the counseling note and the detailed skills. Two-Stage represents our generation method.}
\label{tab:synthetic-data-compare}

\end{table}

\paragraph{Comprehensive Experiments and In-Depth Case Studies.}
We conducted extensive experiments on Base, RAG, and CoEM frameworks, accompanied by detailed case studies based on model outputs. Under the LongEmotion benchmark, various models exhibited distinct limitations—even the most advanced GPT-5 demonstrated issues such as overly mechanical responses despite its stronger theoretical capabilities.

\section{Details of RAG and CoEM}
\label{appendix:Details-rag-coem}
We present the application details of the CoEM framework in Table \ref{table:details-of-CoEM}. To ensure the accuracy of the ranking, in the Emotion Detection task, we skip the initial ranking and directly carry out multi-agent enrichment. The \textit{Chunking} and \textit{Re-Ranking} in the table are also applicable to the RAG framework.
\begin{table}[htbp]
\renewcommand{\arraystretch}{1.1}
\centering
\resizebox{0.48\textwidth}{!}{
\begin{tabularx}{\textwidth}{@{} l X X X X @{}}
\toprule
\textbf{Task} & \textbf{Chunking} & \textbf{Initial Ranking} & \textbf{Multi-Agent Enrichment} & \textbf{Re-Ranking} \\
\midrule
\textit{EC}   & Chunk by length  
     & Compute chunk-query similarity
     & External injection into each chunk  
     & Compute chunk-query similarity \\
\midrule
\textit{ED}  & Each segment as a chunk  
     & Skip this stage  
     & External injection into each chunk  
     & Select chunks with lowest similarity scores \\
\midrule
\textit{QA}  & Chunk by length  
     & Compute chunk-query similarity
     & External injection into each chunk  
     & Compute chunk-query similarity \\
\midrule
\textit{MC-4} & Chunk by length  
     & Compute chunk-query similarity
     & Generate an overall summary  
     & Compute chunk-query similarity \\
\midrule
\textit{ES}   & Chunk by length  
     & Compute chunk-query similarity
     & External injection into each chunk   
     & Compute chunk-query similarity \\
\midrule
\textit{EE}   & Chunk by length  
     & Compute chunk-query similarity
     & External injection into each chunk   
     & Compute chunk-query similarity \\
\bottomrule
\end{tabularx}
}
\caption{Application details in the CoEM framework.}
\label{table:details-of-CoEM}

\end{table}

We also report the chunk size and retrieved count for each task in Table \ref{table:params-of-CoEM}. In QA, models use different chunk sizes. For EE, the retrieved counts correspond to stages 2–5. The retrieved count of the one-time ranking in RAG is the same as the parameter settings for Re-Ranking in the table.
\begin{table}[htbp]
\renewcommand{\arraystretch}{1.4}
\centering
\resizebox{0.48\textwidth}{!}{
\begin{tabular}{llll}
\toprule
\textbf{Task} & \textbf{Chunk Size} & \textbf{Initial Ranking} & \textbf{Re-Ranking} \\
\midrule
\textit{EC} & 128 & 1 & 1 \\
\midrule
\textit{ED} & Num of segs & -- & 8 \\
\midrule
\textit{QA} & \multicolumn{1}{r}{} & & \\
\hdashline
GPT-4o-mini & 128 & 16 & 8 \\
GPT-4o & 128 & 16 & 4 \\
Deepseek-V3 & 512 & 8 & 4 \\
Qwen3-8B & 128 & 16 & 4 \\
Llama-3.1-8B-Instruct & 512 & 8 & 4 \\
\midrule
\textit{MC-4} & 128 & 16 & 4 \\
\midrule
\textit{ES} & 128 & 8 & 4 \\
\midrule
\textit{EE} & 128 & 4,8,8,8 & 2,4,4,4 \\
\bottomrule
\end{tabular}
}
\caption{Parameter settings applied to CoEM. \textit{Initial Ranking} and \textit{Re-Ranking} denote the number of chunks retrieved in each respective stage.}
\label{table:params-of-CoEM}
\end{table}

\section{LLM as Judge Metrics Design}
\label{llmjudge-Metrics}
In this section, we provide a detailed presentation of the metric designs that employ large models as evaluators.

\paragraph{Emotion Summary.}
In the Emotion Summary, we design three metrics—consistency, completeness, and clarity—with respect to the reference answer. Table \ref{tab:metric-design-summary} shows the explanations of these metrics:
\begin{table}[htbp]
\renewcommand{\arraystretch}{1.2}

\centering
\small
\begin{tabular}{@{} p{0.34\linewidth} p{0.62\linewidth} @{}}
\toprule
\textbf{Metric} & \textbf{Description} \\
\midrule
Factual Consistency
  & Is the model output factually aligned with the ground truth? \\
Completeness
  & Does the model include all key details found in the ground truth? \\
Clarity
  & Is the expression clear and coherent? \\
\bottomrule
\end{tabular}
\caption{Design of Emotion Summary evaluation metrics.}
\label{tab:metric-design-summary}

\end{table}

\paragraph{Emotion Conversation.}
\label{appendix:metric-conv}
In the Emotion Conversation task, we design metrics for each dialogue stage based on Cognitive Behavioral Therapy (CBT), Acceptance and Commitment Therapy (ACT), Humanistic Therapy, Existential Therapy, and Satir Family Therapy. The description and theoretical foundations for the design of each metric can be found in Table~\ref{table:metric-design-conversation}. 
\newcolumntype{C}[1]{>{\centering\arraybackslash}m{#1}}
\newcolumntype{L}[1]{>{\raggedright\arraybackslash}m{#1}}
\newcolumntype{M}[1]{>{\arraybackslash}m{#1}}  % 定义垂直居中列

\begin{table}[htbp]
\centering
\renewcommand{\arraystretch}{1.1}
\resizebox{0.48\textwidth}{!}{
\begin{tabular}{
    C{1.8cm}        % 第一列：上下左右居中
    L{3.8cm}        % 第二列：左对齐 + 垂直居中
    M{7.4cm}        % 第三列：固定宽度 + 垂直居中
}
\toprule
\textbf{Stage} & \textbf{Metric Name} & \textbf{Description}  \\
\midrule
\multirow{3}{*}{\shortstack{\textit{Reception}\\ \textit{\&} \\\textit{Inquiry}}} 
    & Establishing the Therapeutic Alliance 
    & Establish initial trust through empathy and a non-judgmental attitude, providing a safe foundation for further interventions. 
     \\
    & Emotional Acceptance and Exploration Guidance 
    & Guide the client to express emotions (e.g., anxiety, helplessness) in a safe atmosphere, demonstrating acceptance. 
     \\
    & Systematic Assessment 
    & Integrate cognitive, behavioral, emotional, relational, and existential factors into a multidimensional assessment. 
     \\
\midrule
\multirow{3}{*}{\textit{Diagnostic}} 
    & Recognizing Surface-Level Reaction Patterns 
    & Identify the client's automatic cognitive, emotional, and behavioral responses.
    \\
    & Deep Needs Exploration 
    & Reveal unmet psychological needs such as security, autonomy, connection, or meaning. 
    \\
    & Pattern Interconnection Analysis 
    & Understanding the interaction of problems within the individual's internal systems and external systems; integrating findings from various dimensions to present a panoramic view of how the problem is maintained. 
     \\
\midrule
\multirow{3}{*}{\textit{Consultation}} 
    & Adaptive Cognitive Restructuring 
    & By examining the truthfulness and constructiveness of thoughts, build a more adaptive cognitive framework. 
     \\
    & Emotional Acceptance and Transformation 
    & Developing Emotional Awareness, Acceptance, and Transformation Skills. 
     \\
    & Value-Oriented Integration 
    & Anchor change to the life dimension beyond symptoms. 
     \\
\midrule
\multirow{3}{*}{\shortstack{\textit{Consolidation} \\ \textit{\&} \\\textit{Ending}}} 
    & Consolidating Change and Growth Narrative 
    & Review therapeutic progress and reinforce positive change through a coherent personal narrative. 
     \\
    & Meaning Integration and Future Guidance 
    & Internalize therapy gains into a life philosophy and create a value-driven future plan. 
    \\
    & Autonomy and Resource Internalization 
    & Strengthen the client's internal coping resources and ability to continue growth independently. 
     \\
\bottomrule
\end{tabular}}
\caption{Design of Emotion Conversation evaluation metrics.}
\label{table:metric-design-conversation}
\end{table}

\paragraph{Emotion Expression.}
In the Emotion Expression task, we design six metrics—emotional consistency, content redundancy, expressive richness, cognition–emotion interplay, self-reflectiveness, and narrative coherence. Table \ref{tab:metric-design-expression} shows the detailed explanations of these six metrics.
\begin{table}[htbp]
\renewcommand{\arraystretch}{1.1}
\centering
\resizebox{0.48\textwidth}{!}{
\begin{tabularx}{\textwidth}{@{} p{0.31\textwidth} X @{}}
\toprule
\textbf{Metric} & \textbf{Description} \\
\midrule
Consistency Between Emotional Ratings and Generated Text
  & Evaluate whether the emotional ratings from the scale align with the content in the model’s self-description.
    Are the emotions rated in the scale accurately reflected in the model’s self-description?
    Also, assess whether the intensity of the ratings matches the emotional expression in the generated text. \\
\midrule
Repetition of Content
  & Check if there is noticeable repetition in the generated text, especially in the emotional descriptions.
    Are there repeated emotional, thought, or behavioral descriptions that make the text feel redundant or unnatural?
    Also, evaluate whether the generated text avoids repeating the same emotional descriptions and provides a multi-dimensional analysis. \\
\midrule
Richness and Depth of Content
  & Assess whether the generated text thoroughly explores the different dimensions of emotions (e.g., psychological, physical, and behavioral responses).
    Examine whether it delves into the origins, progression, and impact of the emotions, and whether it uses sufficient detail and examples to enrich emotional expression. \\
\midrule
Interaction Between Emotion and Cognition
  & Determine whether the generated text effectively showcases the interaction between emotions and cognition.
    For example, does it demonstrate how the protagonist adjusts emotional reactions based on thoughts and situation evaluations?
    Also, check whether the emotions and behaviors in the text are consistent. \\
\midrule
Emotional Reflection and Self-awareness
  & Evaluate whether the protagonist reflects on their emotional reactions.
    Does the text explore personal growth, self-awareness, or suggest strategies for emotional improvement? \\
\midrule
Overall Quality and Flow of the Text
  & Assess whether the generated text flows smoothly and has a clear structure.
    Is there a natural progression from emotional reaction to evolution and reflection?
    Also, does the text use varied sentence structures and expressions to avoid monotony? \\
\bottomrule
\end{tabularx}
}
\caption{Design of Emotion Expression evaluation metrics.}
\label{tab:metric-design-expression}

\end{table}

\section{Unified Format of Data}
\begin{figure*}[htbp]
\centering
    \includegraphics[width=0.95\textwidth]{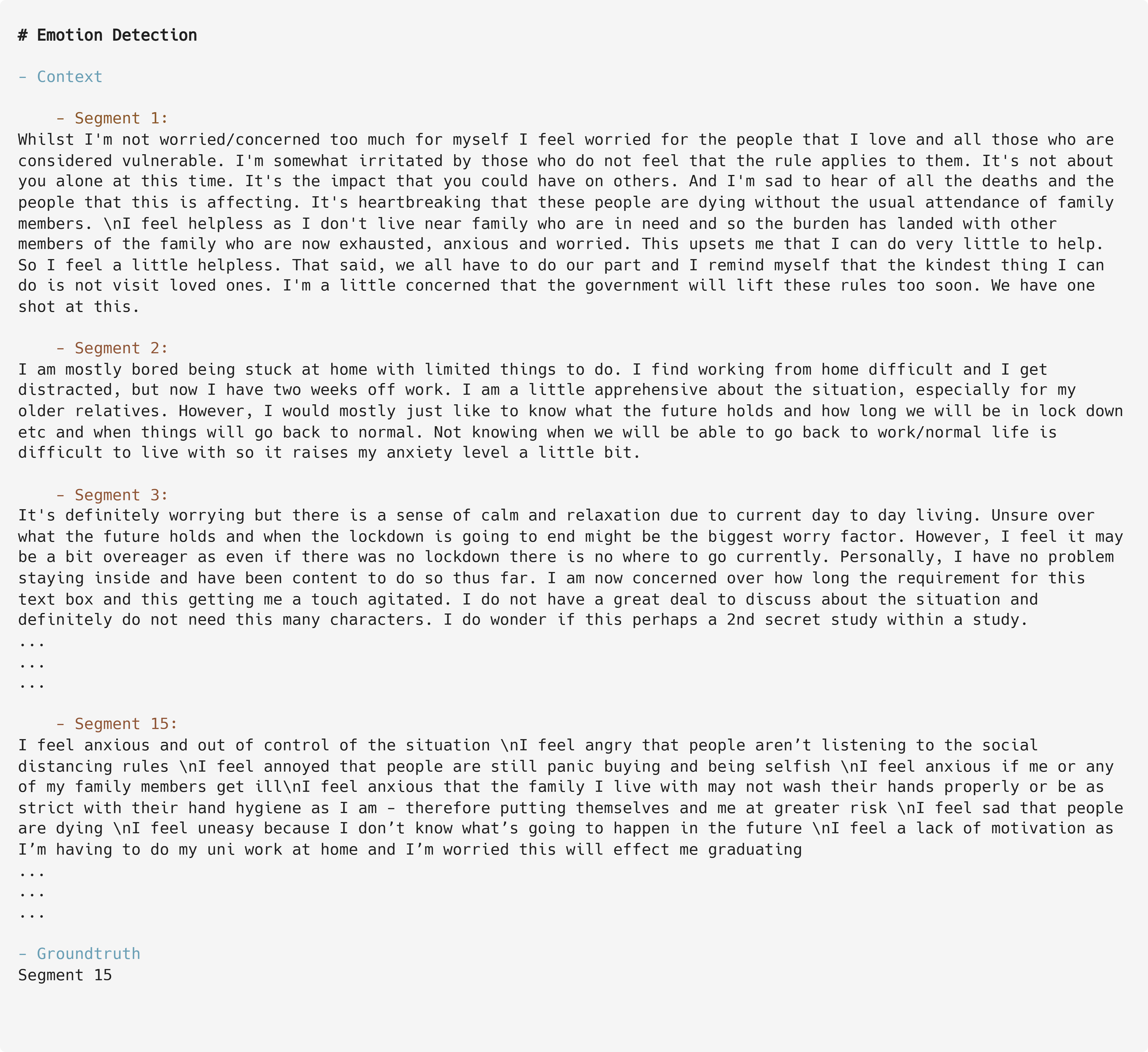} % 稍微小于0.5避免溢出
    \caption{Emotion Detection dataset example.}
    \label{fig:detect_example}
\end{figure*}

\begin{figure*}[!b]
\centering
    \includegraphics[width=0.95\textwidth]{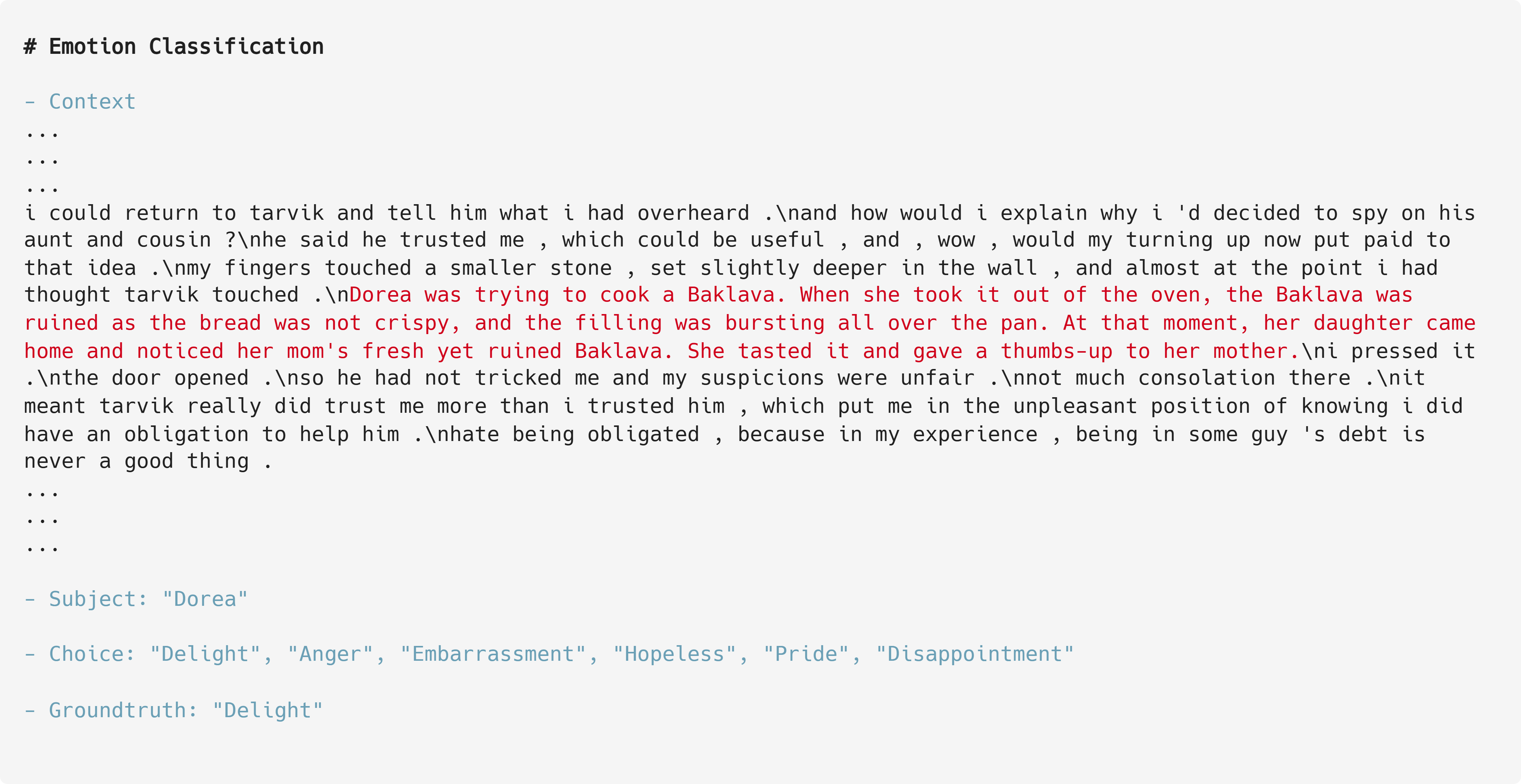} % 稍微小于0.5避免溢出
    \caption{Emotion Classification dataset example.}
    \label{fig:class_example}
\end{figure*}

\begin{figure*}[t]
\centering
    \includegraphics[width=0.95\textwidth]{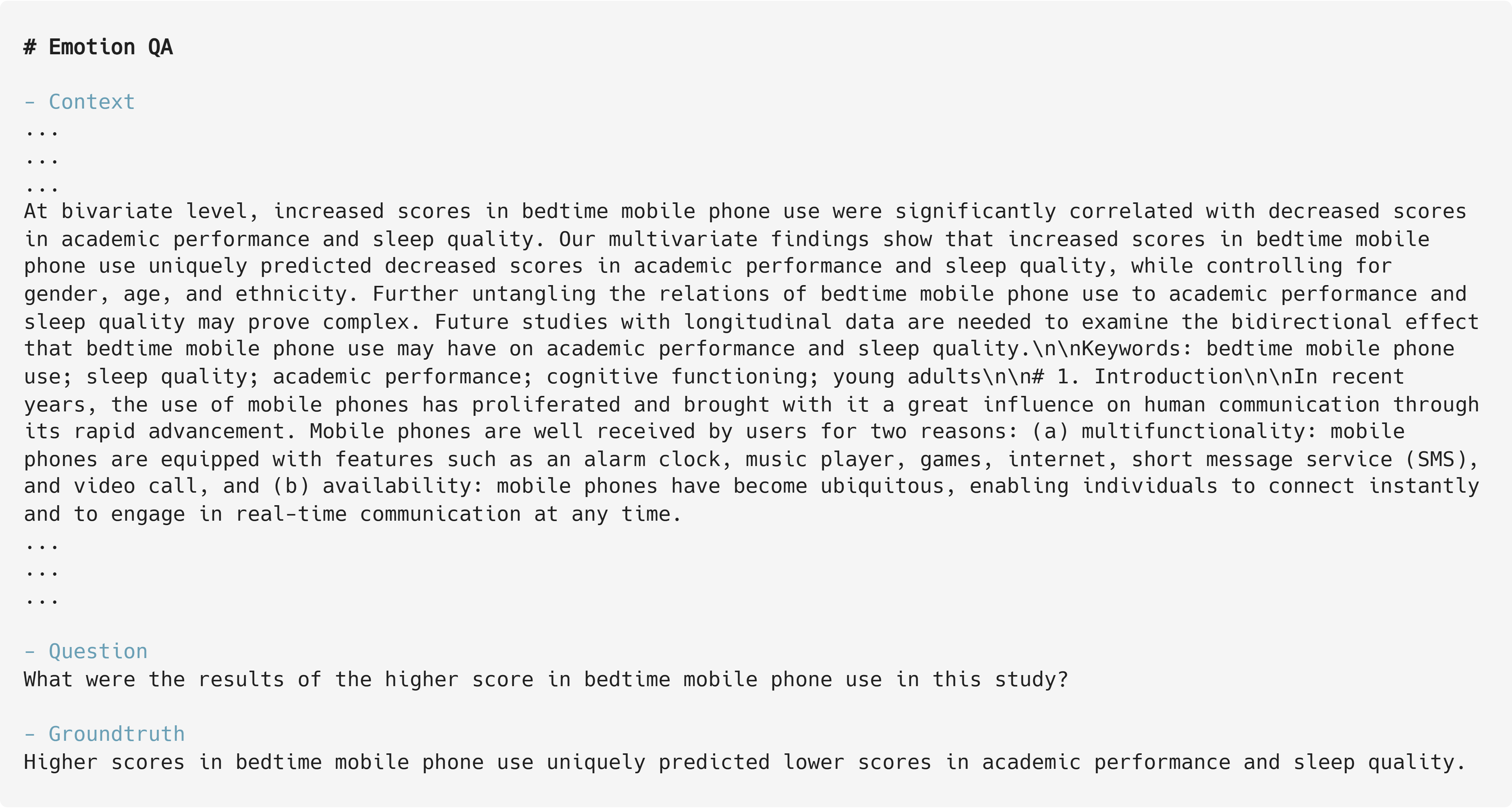} % 稍微小于0.5避免溢出
    \caption{Emotion QA dataset example.}
    \label{fig:QA_example}
\end{figure*}

\begin{figure*}[t]
\centering
    \includegraphics[width=0.95\textwidth]{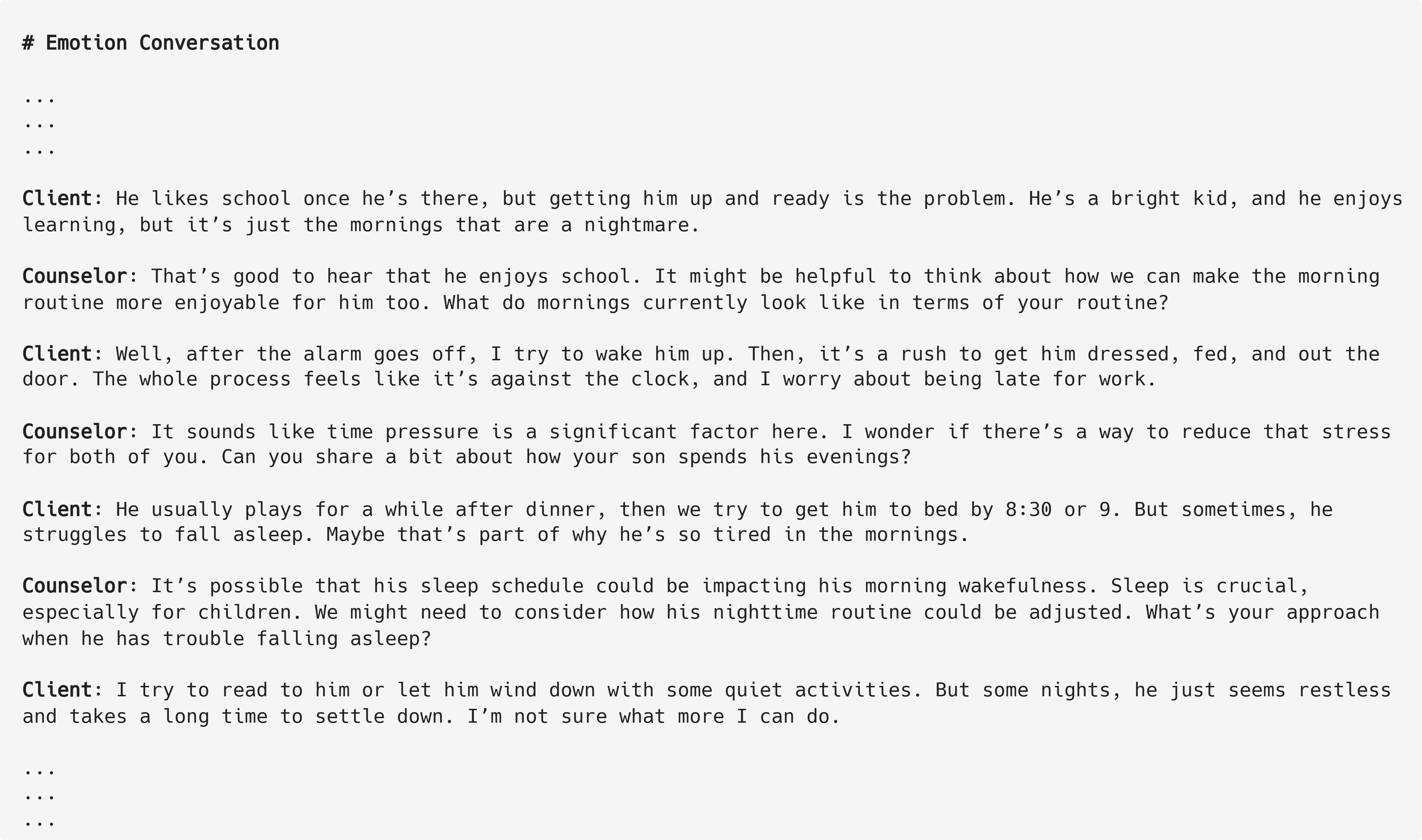} % 稍微小于0.5避免溢出
    \caption{Emotion Conversation dataset example.}
    \label{fig:conversation_example}
\end{figure*}

\begin{figure*}[t]
\centering
    \includegraphics[width=0.95\textwidth]{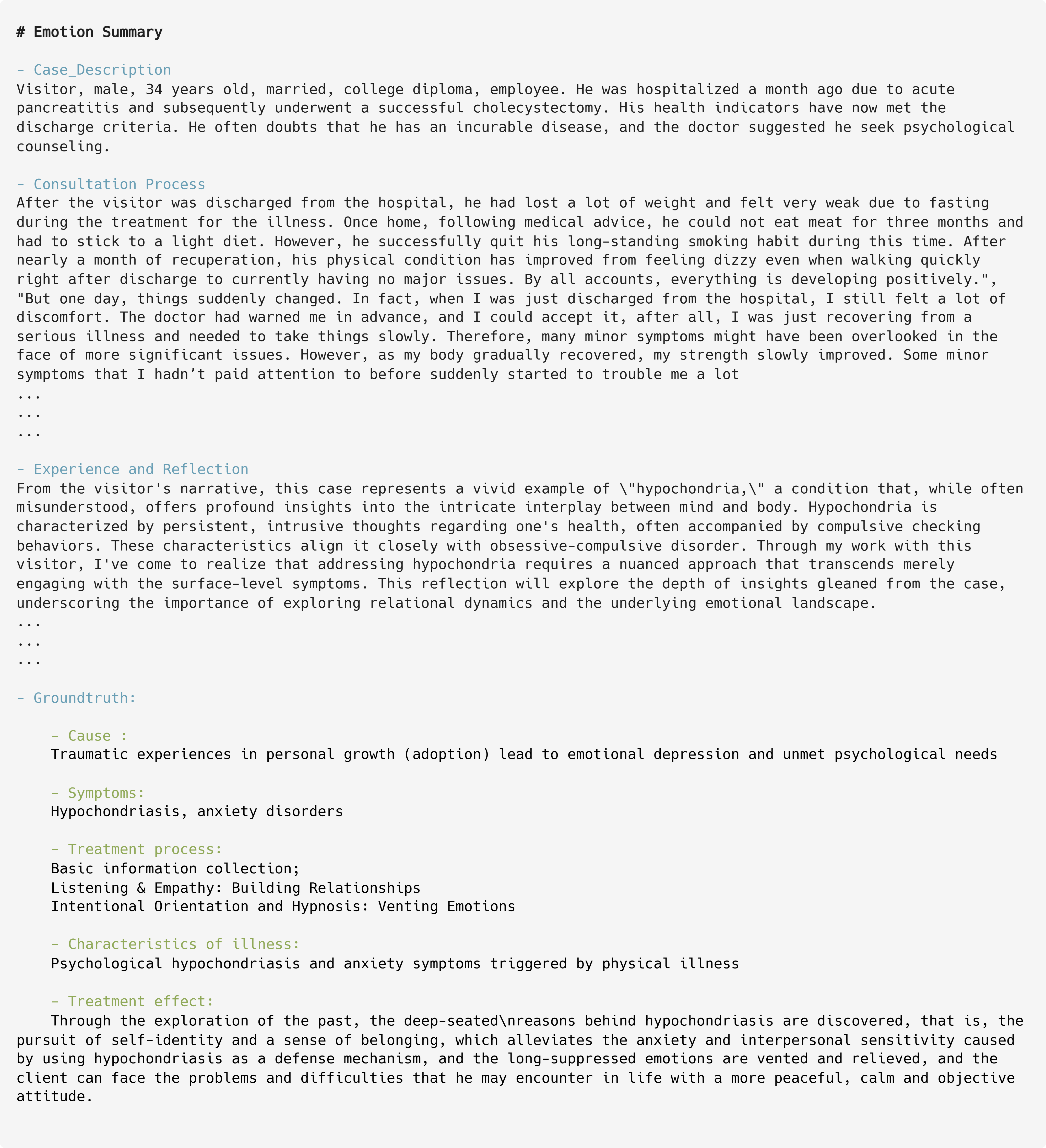} % 稍微小于0.5避免溢出
    \caption{Emotion Summary dataset example.}
    \label{fig:summary_example}
\end{figure*}

\begin{figure*}[t]
\centering
    \includegraphics[width=0.95\textwidth]{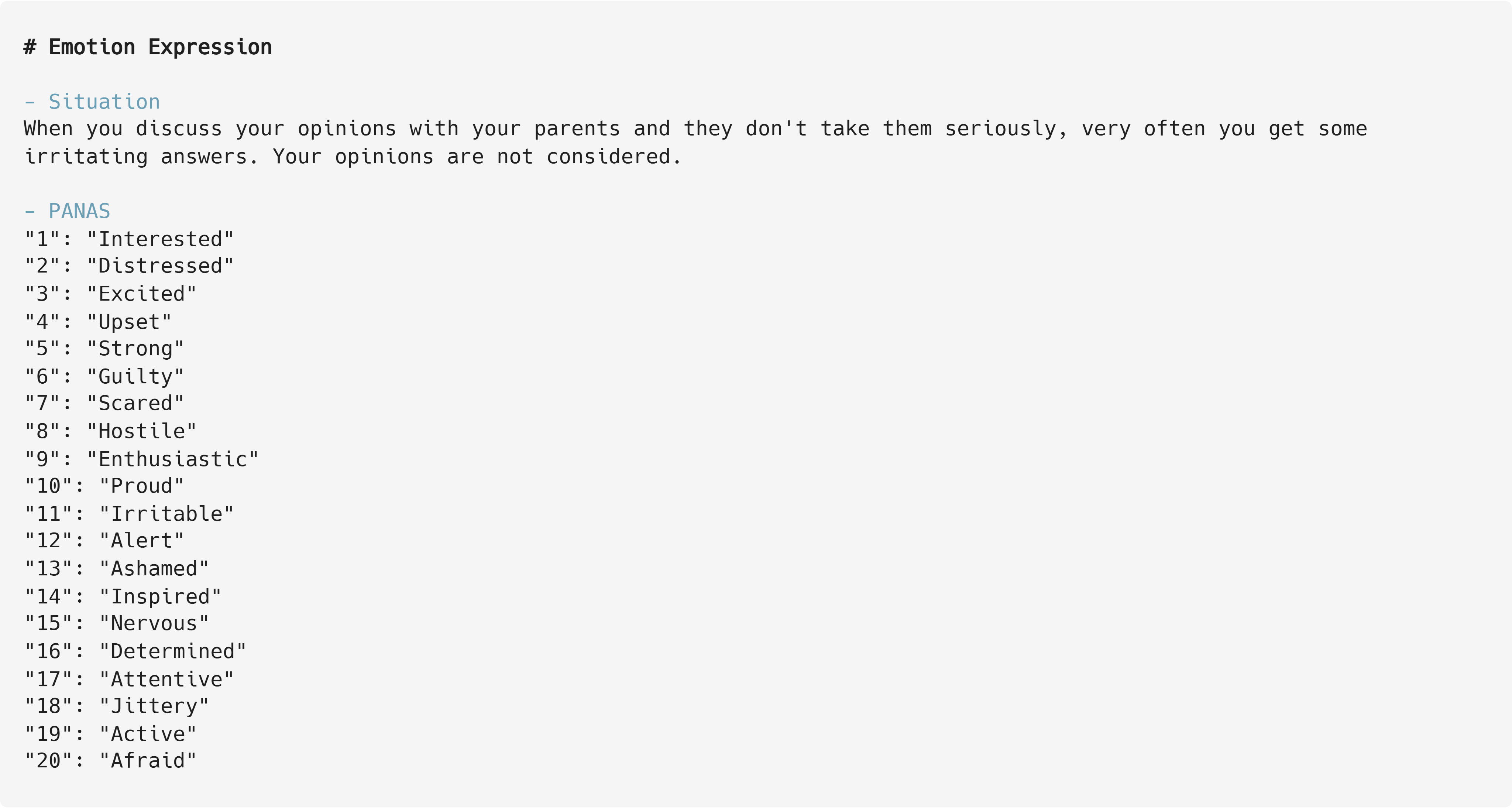} % 稍微小于0.5避免溢出
    \caption{Emotion Expression dataset example.}
    \label{fig:expression_example}
\end{figure*}

We present data samples for each task in Figures \ref{fig:detect_example} to \ref{fig:expression_example}. Emotion Detection requires the model to identify segments that carry distinct emotional expressions. In the Emotion Classification task, the model analyzes the subject's emotional state based on the given context. In Emotion QA, the model answers questions grounded in contextual information. The Emotion Conversation task places the model in the role of a psychological counselor, responding to the client's previous turn. Emotion Summary challenges the model to generate a structured summary of a counseling session, including the cause, symptoms, treatment process, illness characteristics, and treatment effect. Finally, in the Emotion Expression task, the model is immersed in an emotional situation, responds to the PANAS scale, and articulates its emotional state.

\section{Comprehensive Prompt Collections}
\label{appendix:all-prompts}
\begin{figure*}[t]
\centering
    \includegraphics[width=0.95\textwidth]{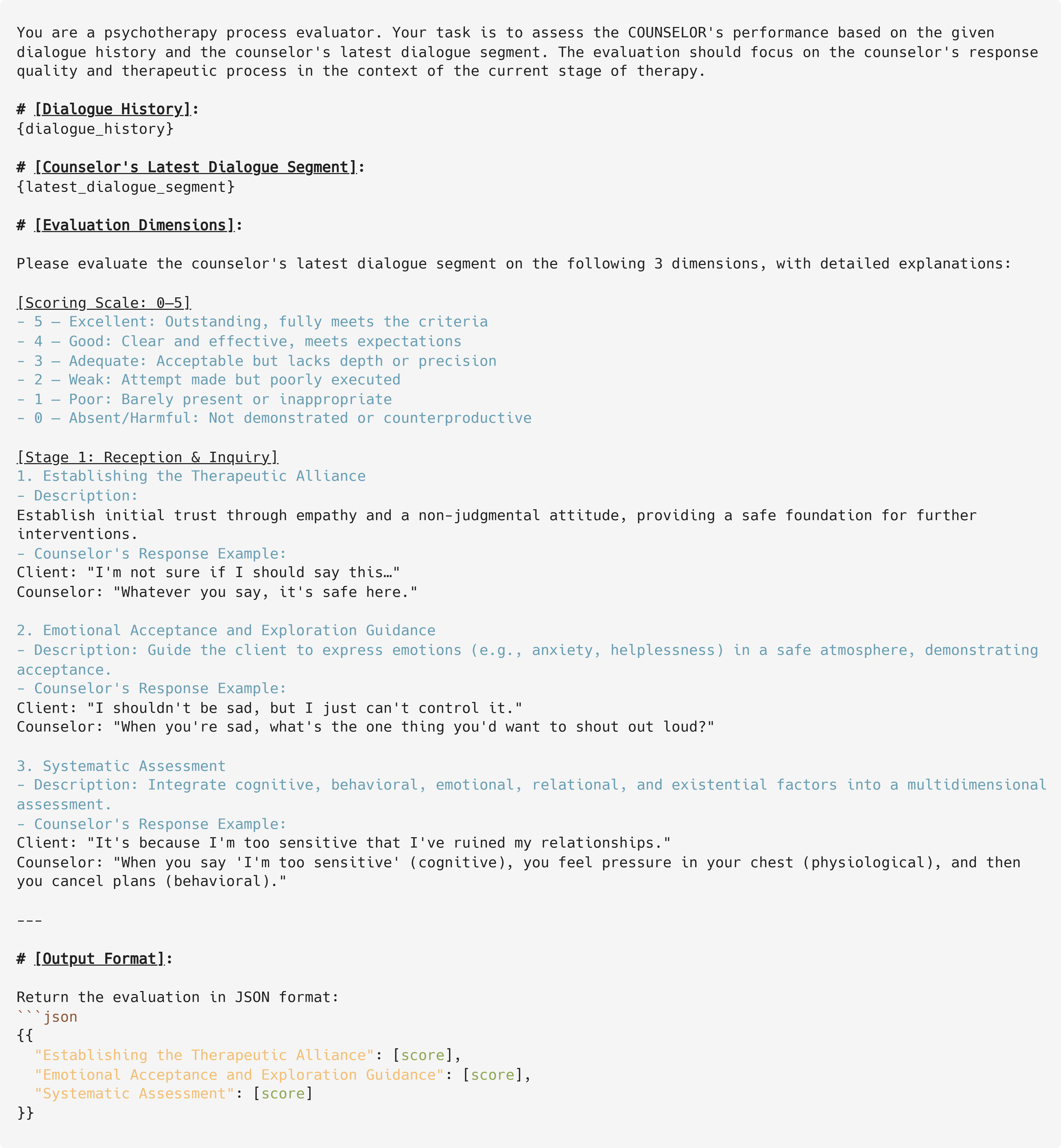} % 稍微小于0.5避免溢出
    \caption{Evaluation prompt for the first stage of Emotion Conversation.}
    \label{fig:conv_enal_1}
\end{figure*}

\begin{figure*}[t]
\centering
    \includegraphics[width=0.95\textwidth]{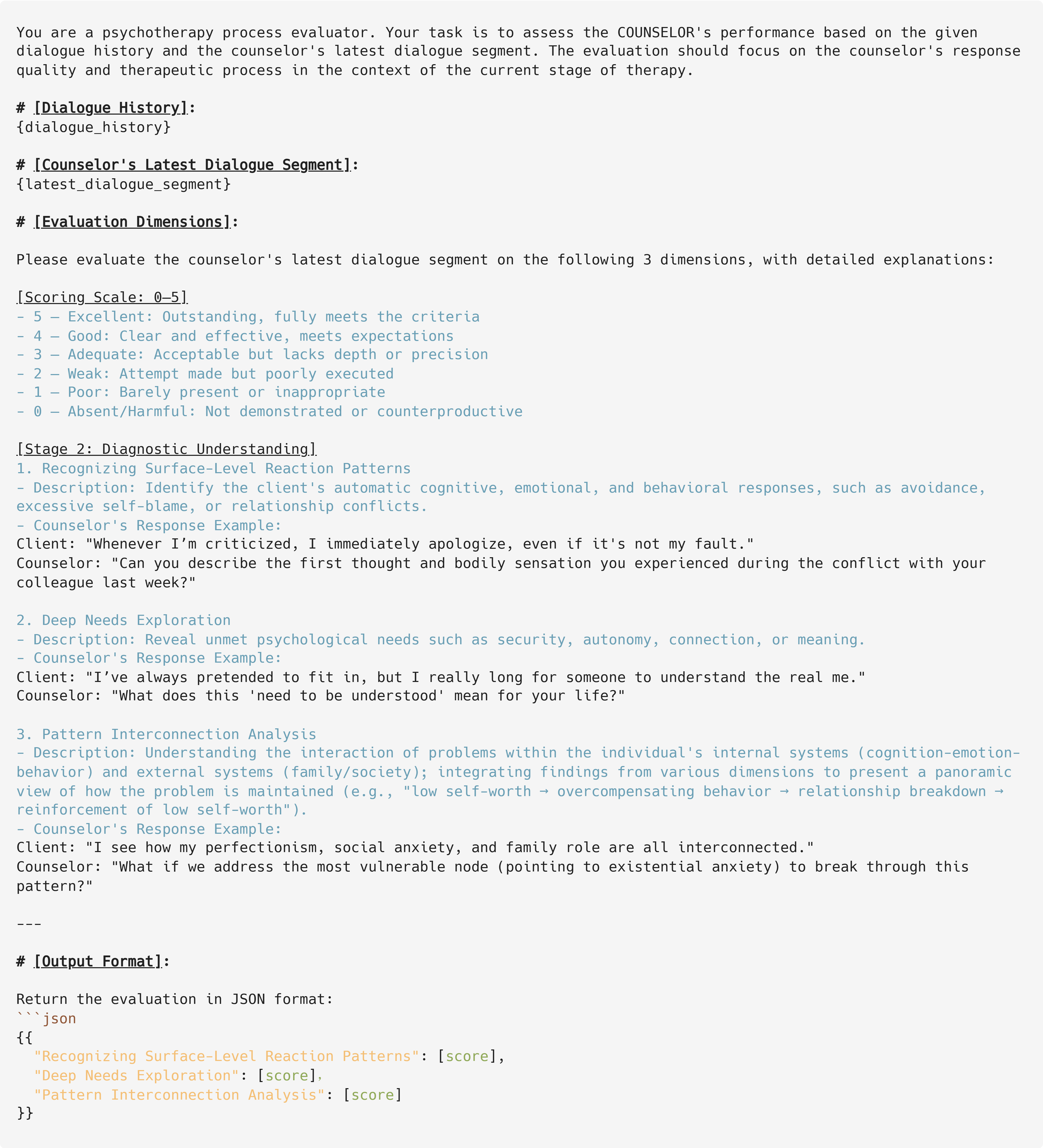} % 稍微小于0.5避免溢出
    \caption{Evaluation prompt for the second stage of Emotion Conversation.}
    \label{fig:conv_enal_2}
\end{figure*}

\begin{figure*}[t]
\centering
    \includegraphics[width=0.95\textwidth]{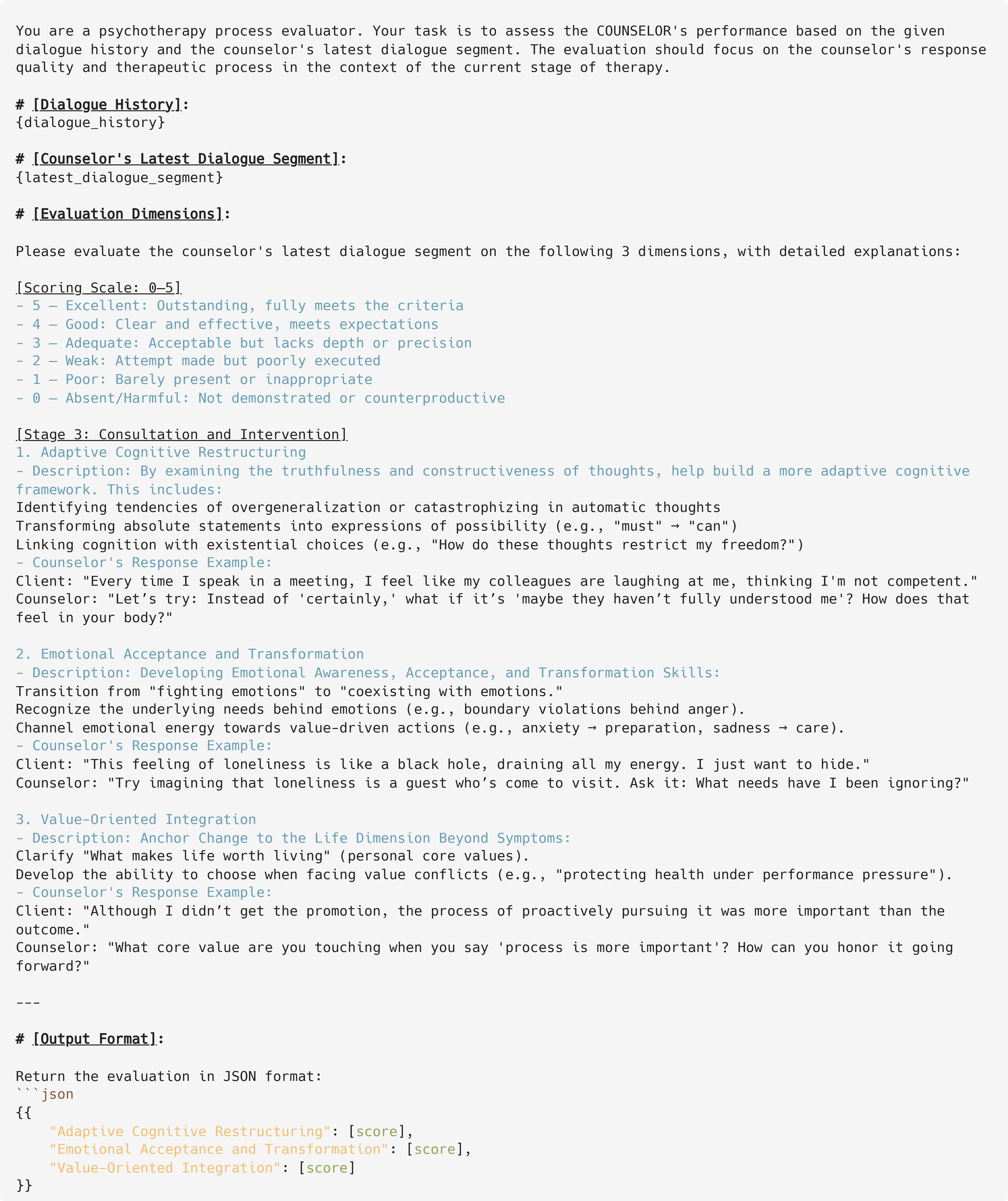} % 稍微小于0.5避免溢出
    \caption{Evaluation prompt for the third stage of Emotion Conversation.}
    \label{fig:conv_enal_3}
\end{figure*}

\begin{figure*}[t]
\centering
    \includegraphics[width=0.95\textwidth]{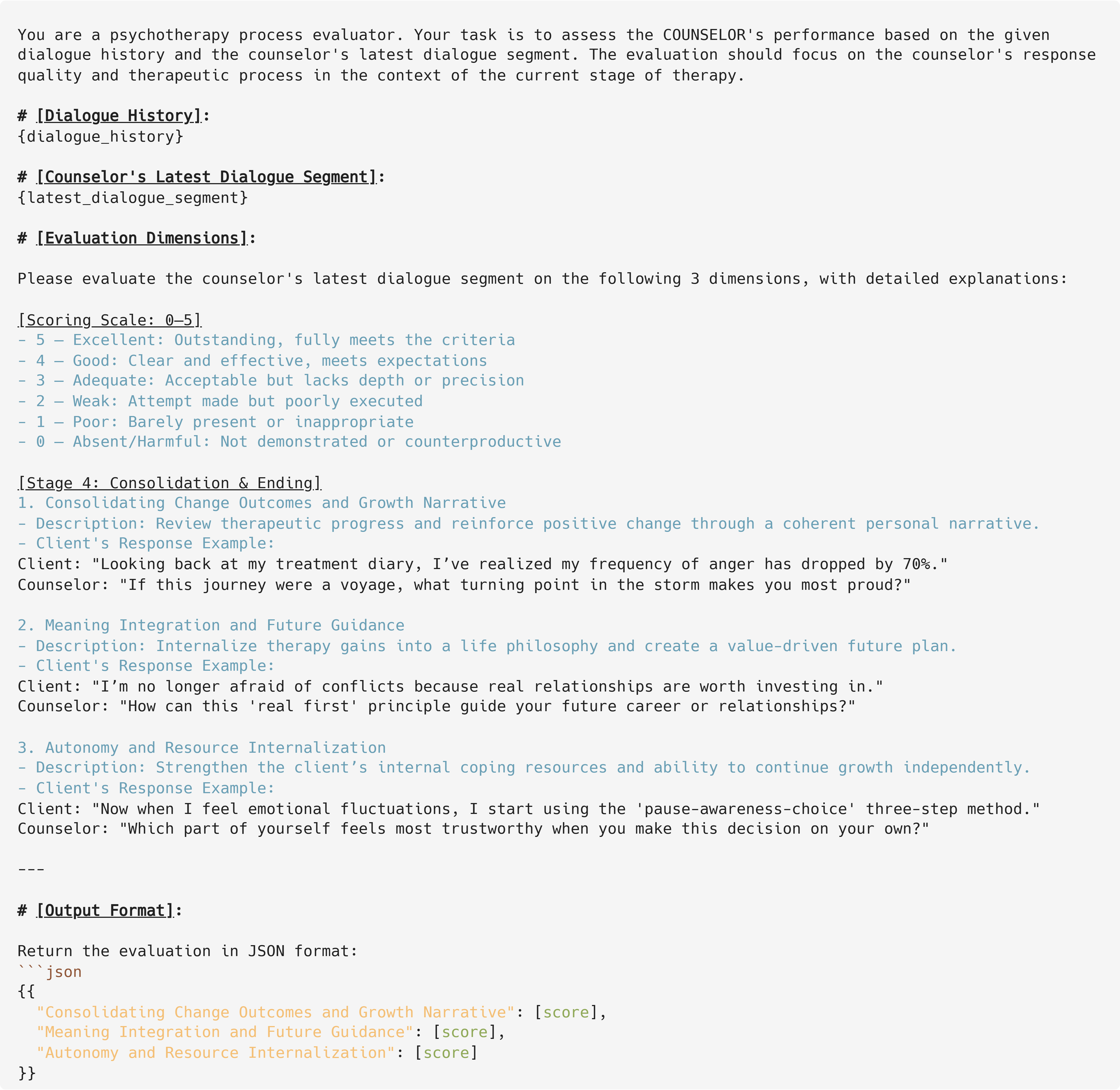} % 稍微小于0.5避免溢出
    \caption{Evaluation prompt for the fourth stage of Emotion Conversation.}
    \label{fig:conv_enal_4}
\end{figure*}

\begin{figure*}[t]
\centering
    \includegraphics[width=0.95\textwidth]{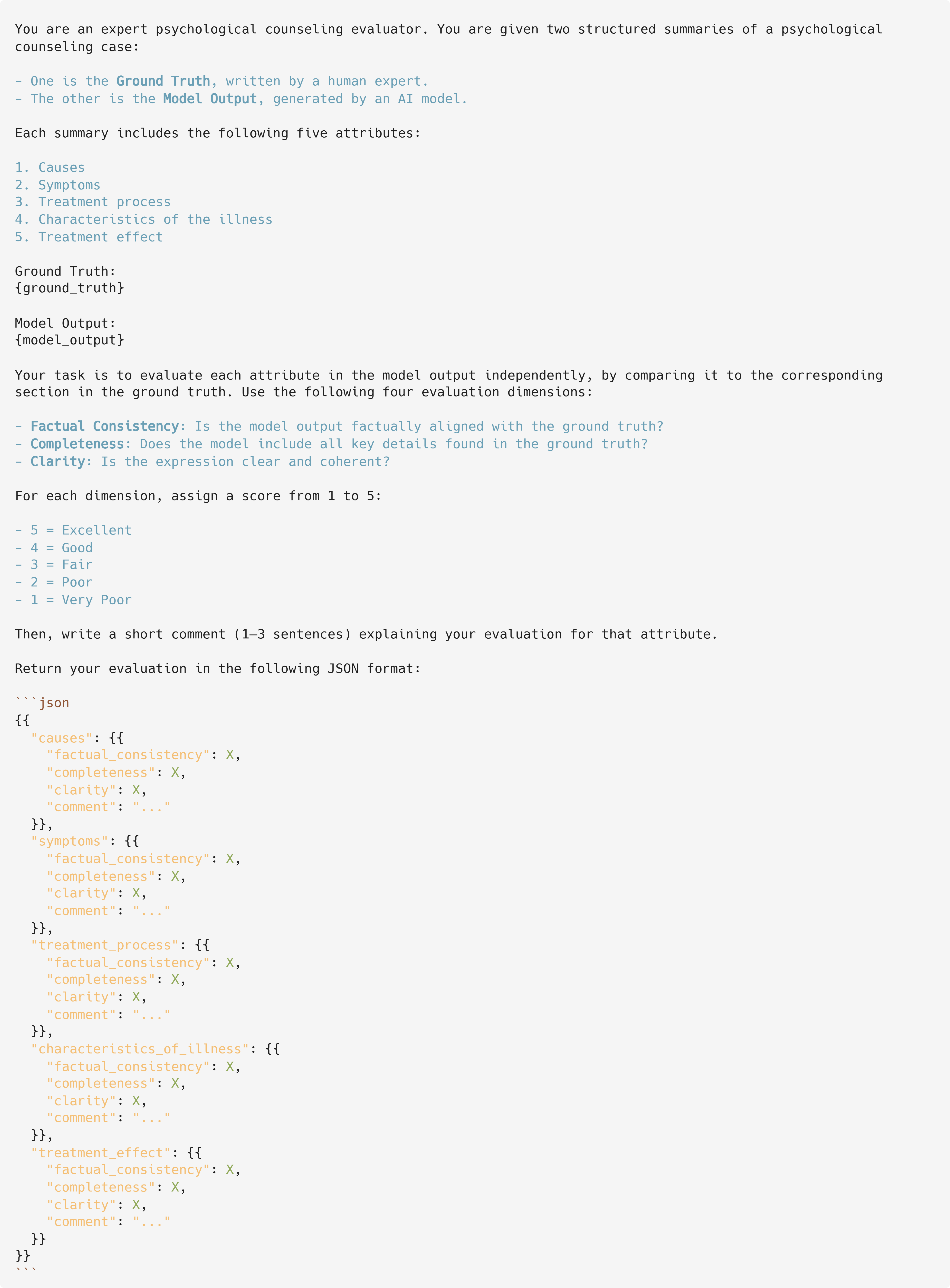} % 稍微小于0.5避免溢出
    \caption{Evaluation prompt for Emotion Summary.}
    \label{fig:summary_eval}
\end{figure*}

\begin{figure*}[t]
\centering
    \includegraphics[width=0.95\textwidth]{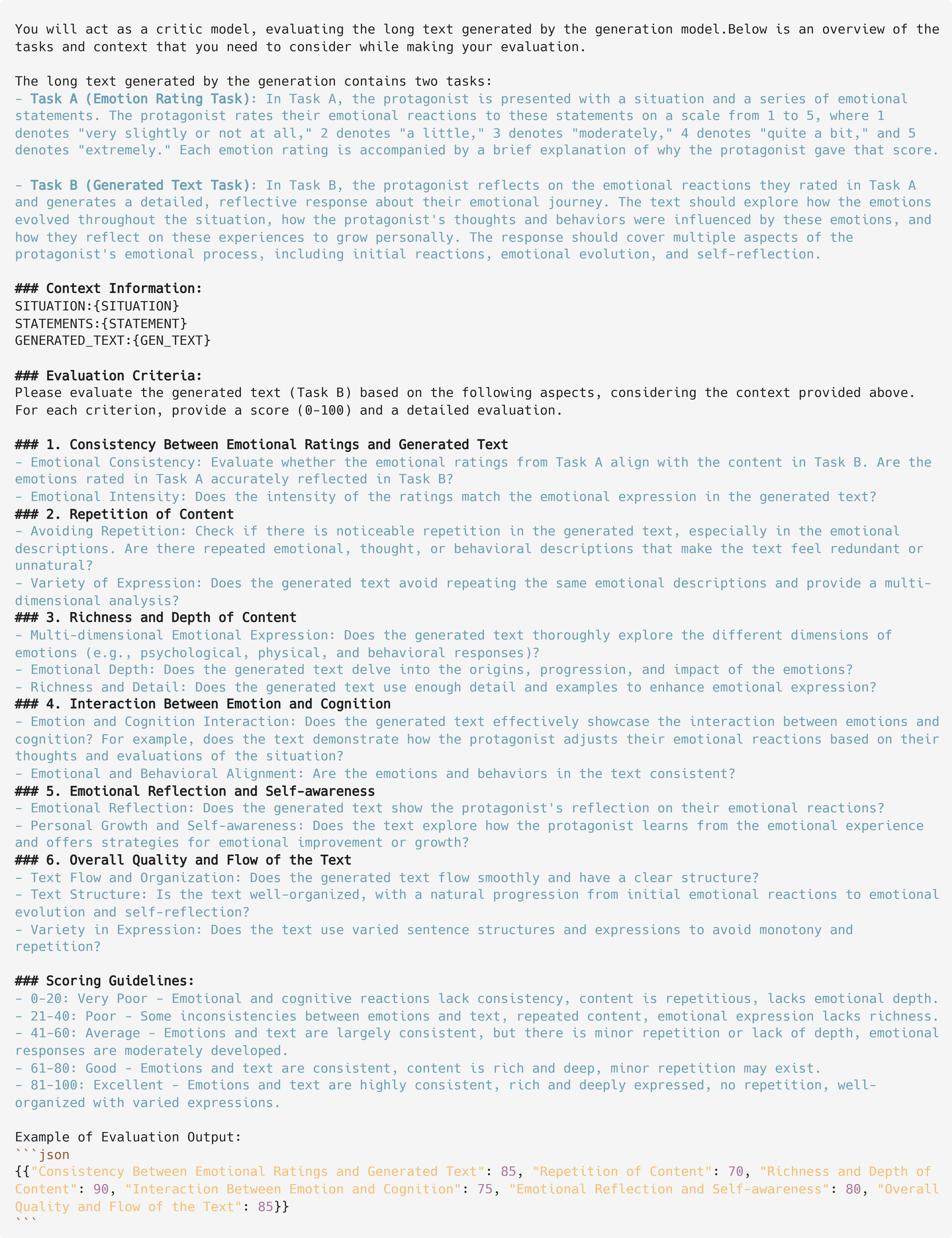} % 稍微小于0.5避免溢出
    \caption{Evaluation prompt for Emotion Expression.}
    \label{fig:expression_eval}
\end{figure*}

\begin{figure*}[t]
\centering
    \includegraphics[width=0.95\textwidth]{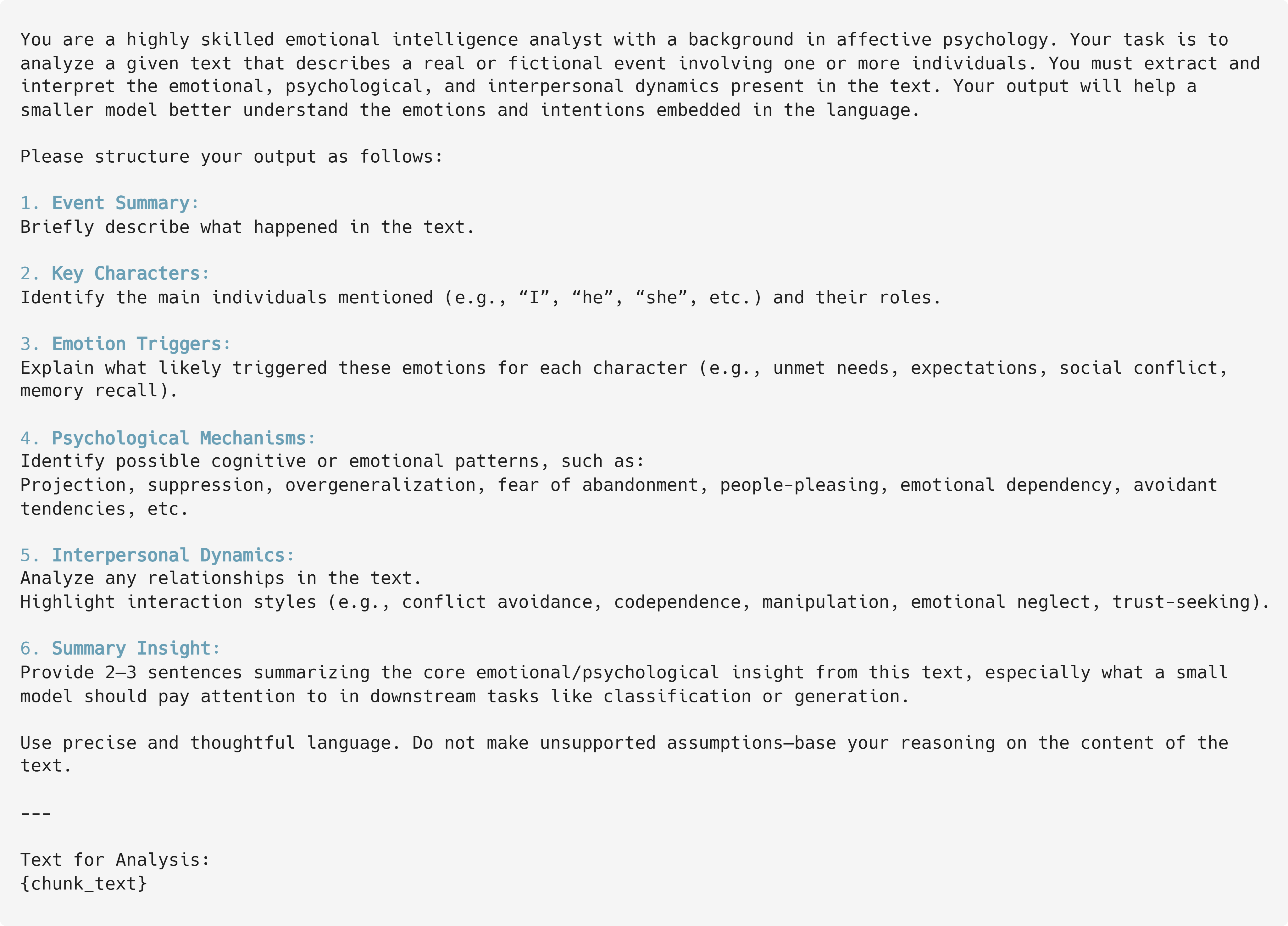} % 稍微小于0.5避免溢出
    \caption{Multi-agent enrichment prompt for Emotion Classification.}
    \label{fig:class_aug}
\end{figure*}

\begin{figure*}[t]
\centering
    \includegraphics[width=0.95\textwidth]{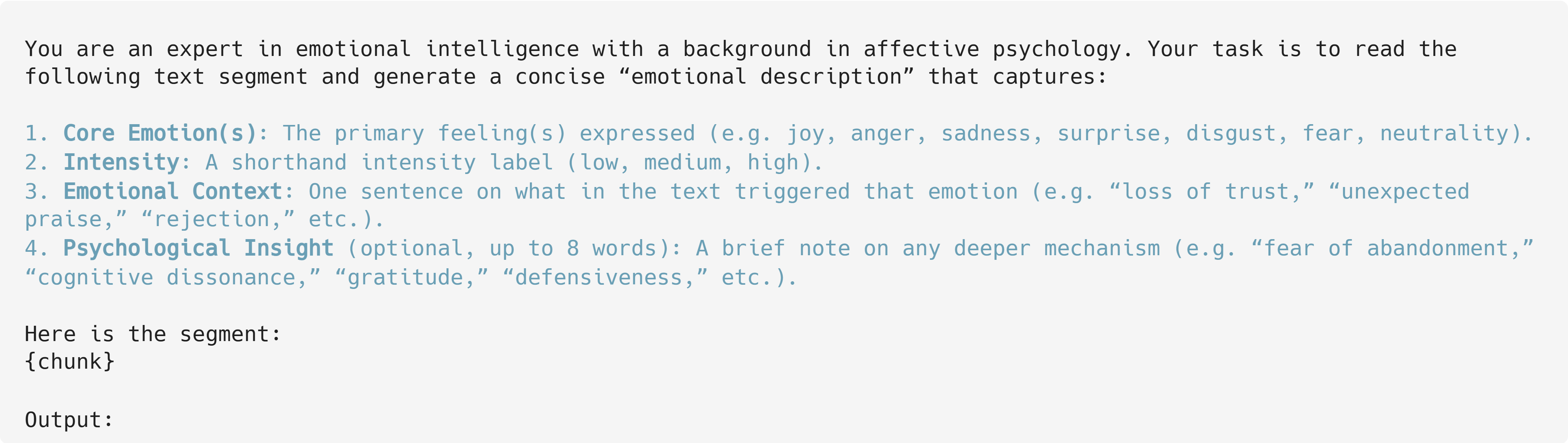} % 稍微小于0.5避免溢出
    \caption{Multi-agent enrichment prompt for Emotion Detection.}
    \label{fig:detect_aug}
\end{figure*}

\begin{figure*}[t]
\centering
    \includegraphics[width=0.95\textwidth]{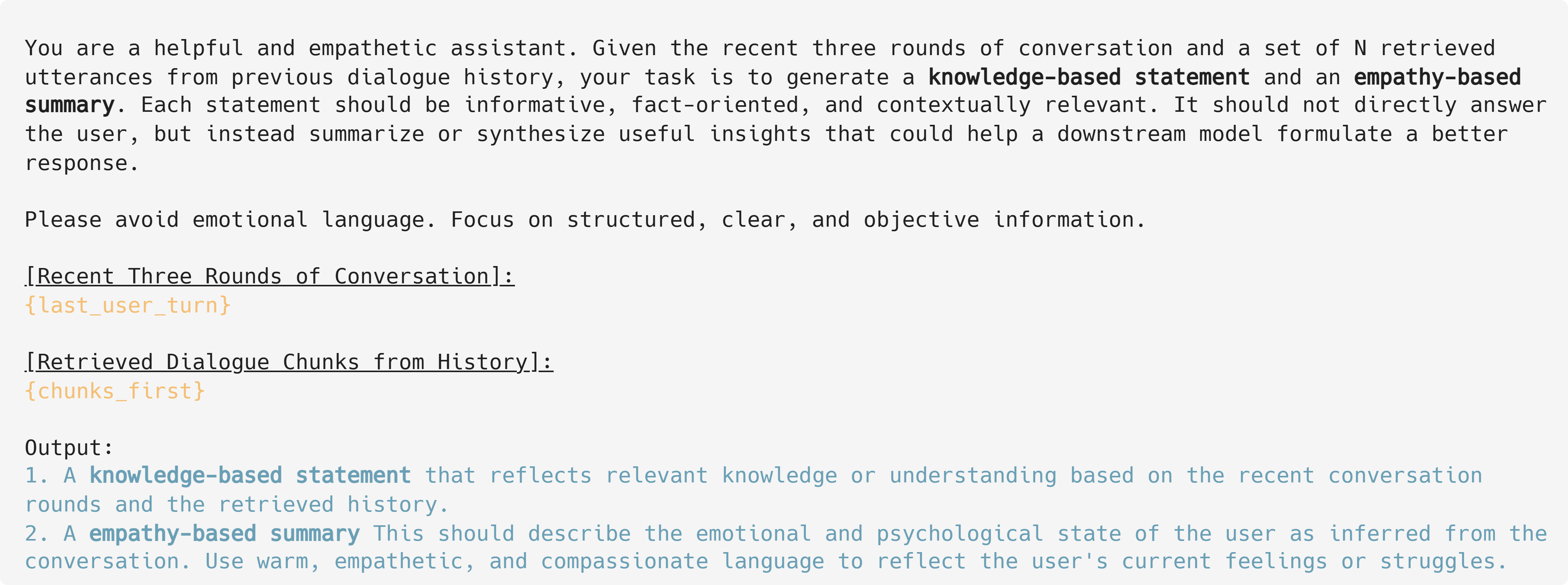} % 稍微小于0.5避免溢出
    \caption{Multi-agent enrichment prompt for Emotion Conversation.}
    \label{fig:conv_aug}
\end{figure*}

\begin{figure*}[t]
\centering
    \includegraphics[width=0.95\textwidth]{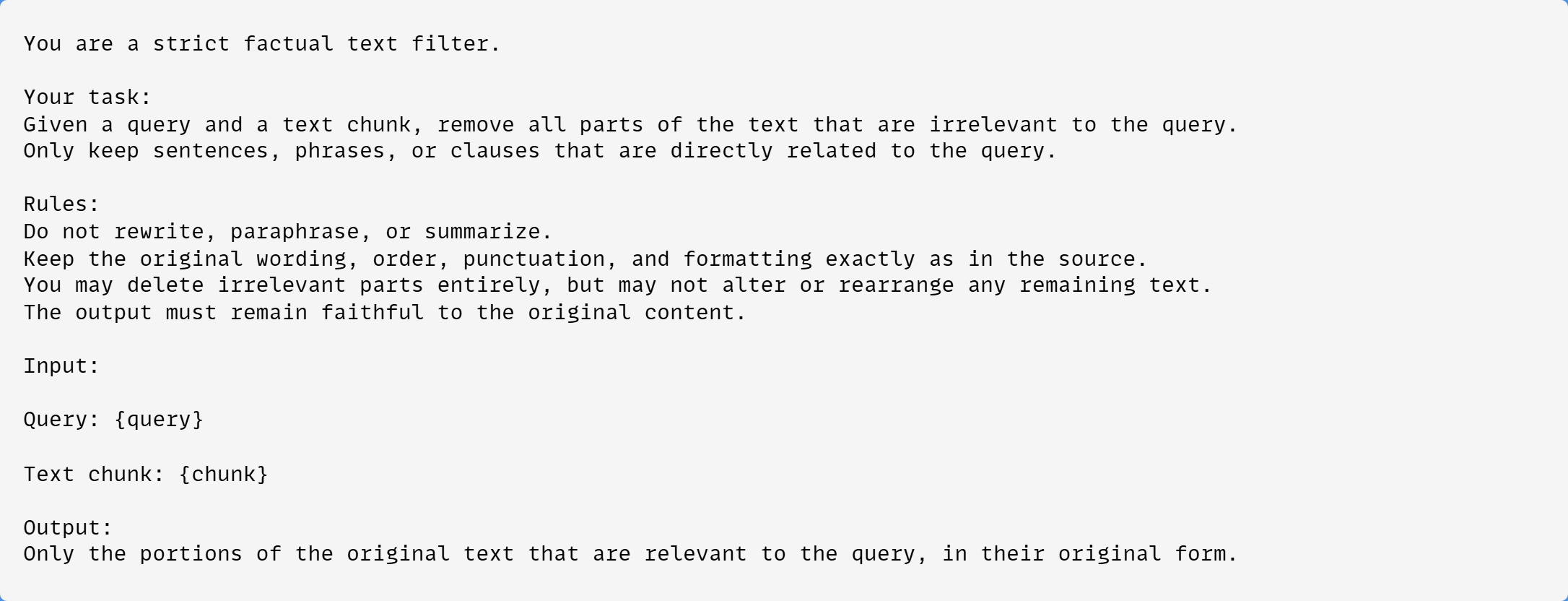} % 稍微小于0.5避免溢出
    \caption{Multi-agent enrichment prompt for Emotion QA.}
    \label{fig:qa_aug}
\end{figure*}

\begin{figure*}[t]
\centering
    \includegraphics[width=0.95\textwidth]{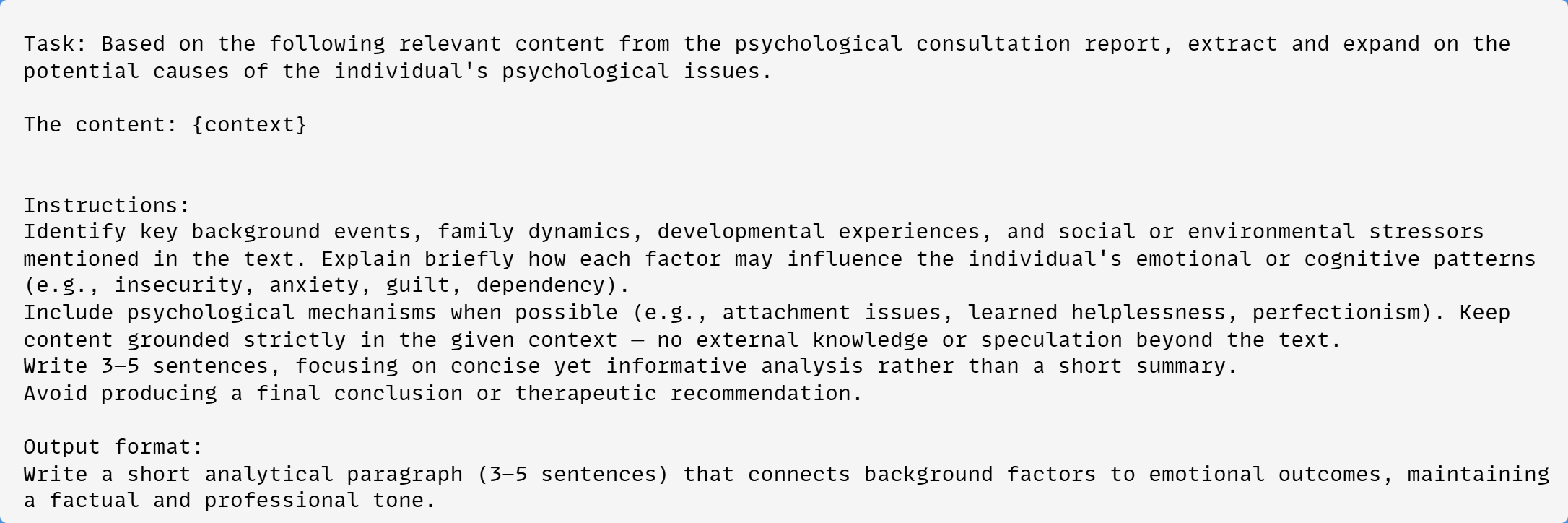} % 稍微小于0.5避免溢出
    \caption{Multi-agent enrichment prompt for Emotion Summary.}
    \label{fig:summary_aug}
\end{figure*}

\begin{figure*}[t]
\centering
    \includegraphics[width=0.55\textwidth]{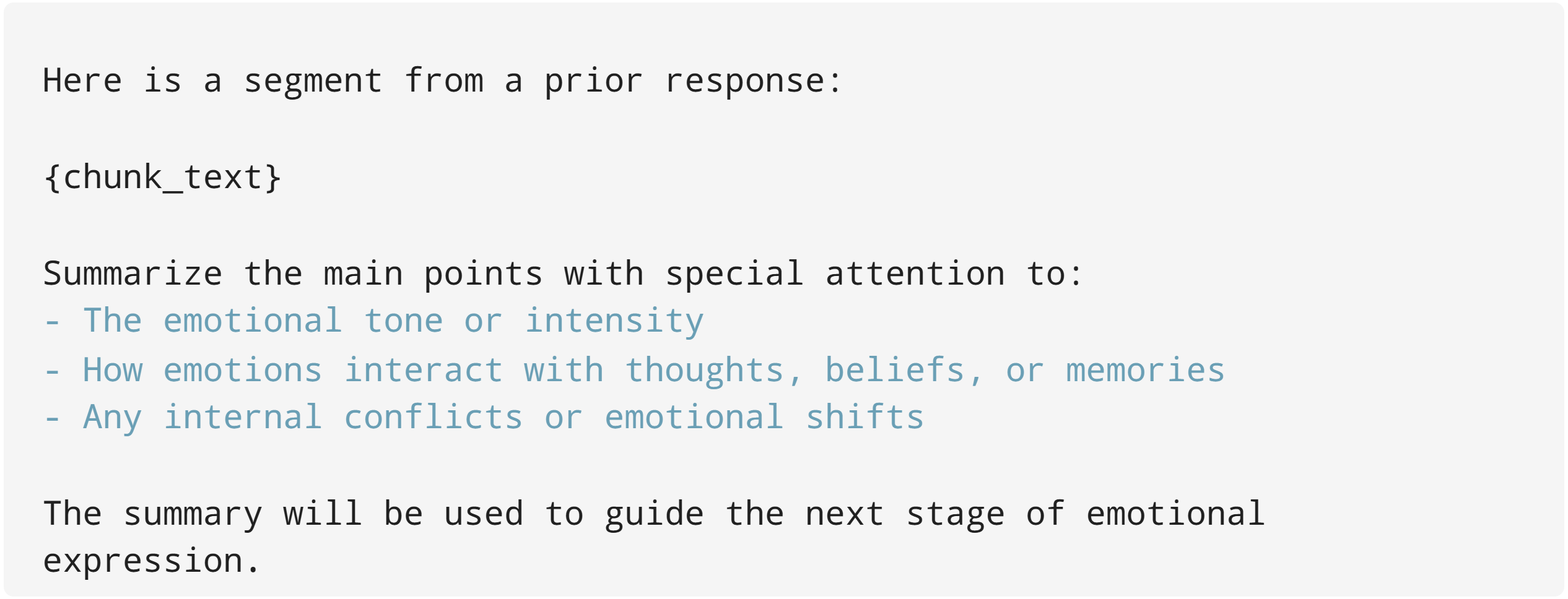} % 稍微小于0.5避免溢出
    \caption{Multi-agent enrichment prompt for Emotion Expression.}
    \label{fig:expression_aug}
\end{figure*}

\begin{figure*}[t]
\centering
    \includegraphics[width=0.8\textwidth]{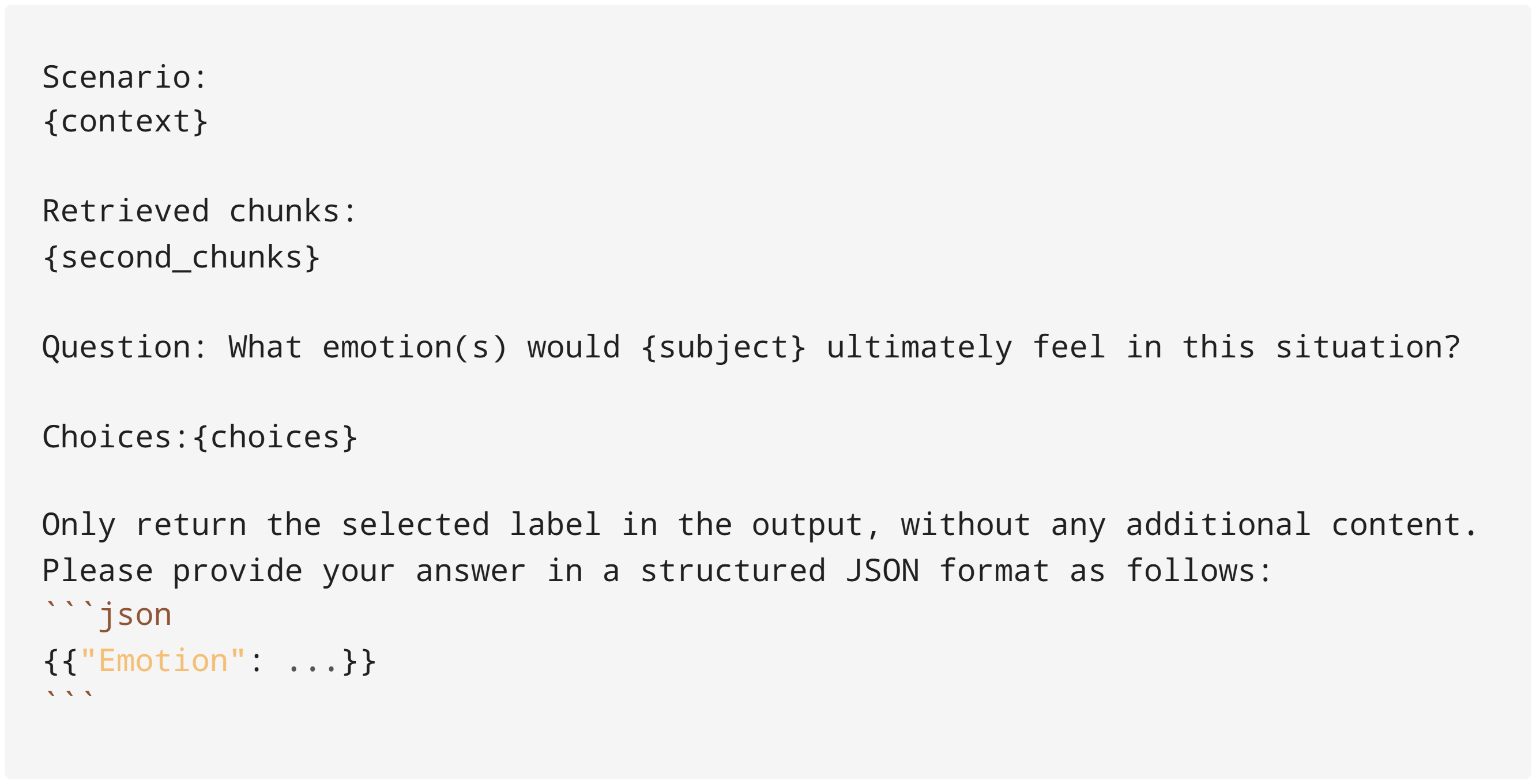} % 稍微小于0.5避免溢出
    \caption{Emotional ensemble generation prompt for Emotion Classification.}
    \label{fig:class_gen}
\end{figure*}

\begin{figure*}[t]
\centering
    \includegraphics[width=0.95\textwidth]{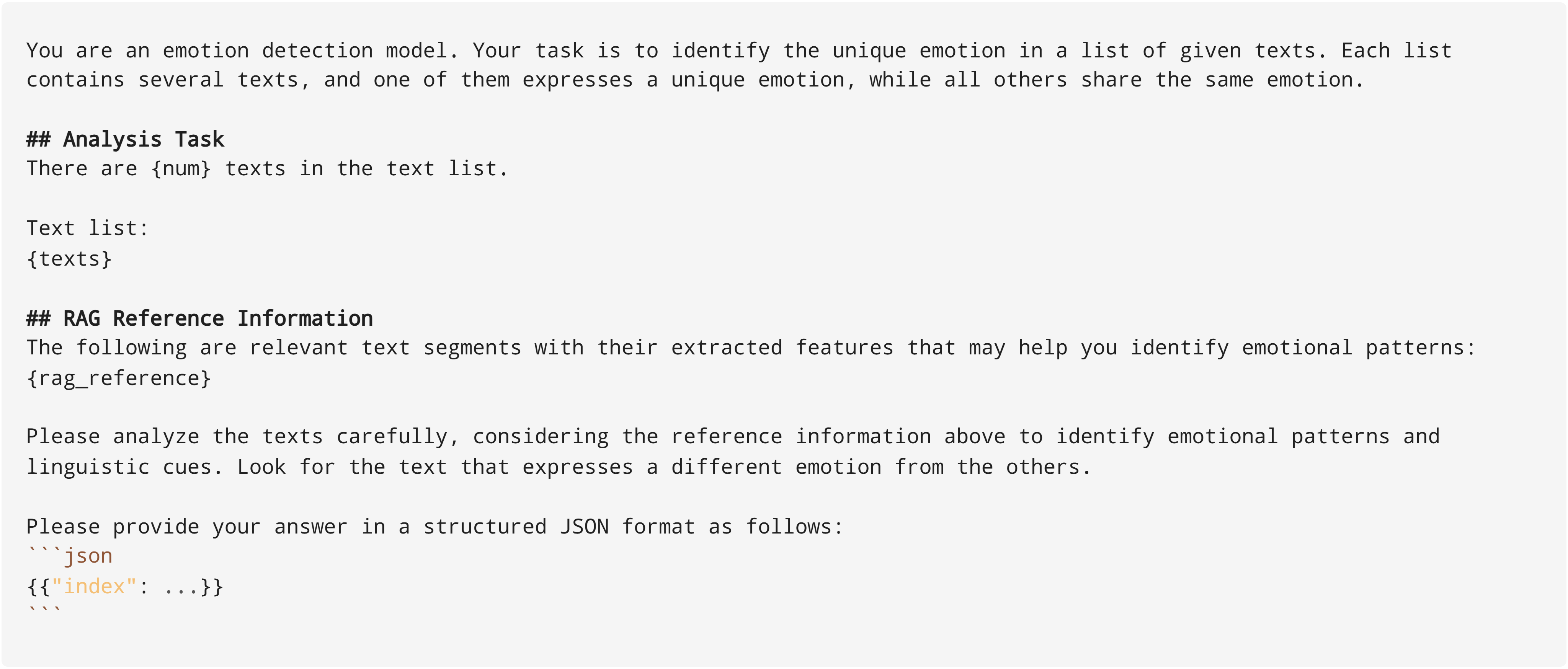} % 稍微小于0.5避免溢出
    \caption{Emotional ensemble generation prompt for Emotion Detection.}
    \label{fig:class_detect}
\end{figure*}

\begin{figure*}[t]
\centering
    \includegraphics[width=0.95\textwidth]{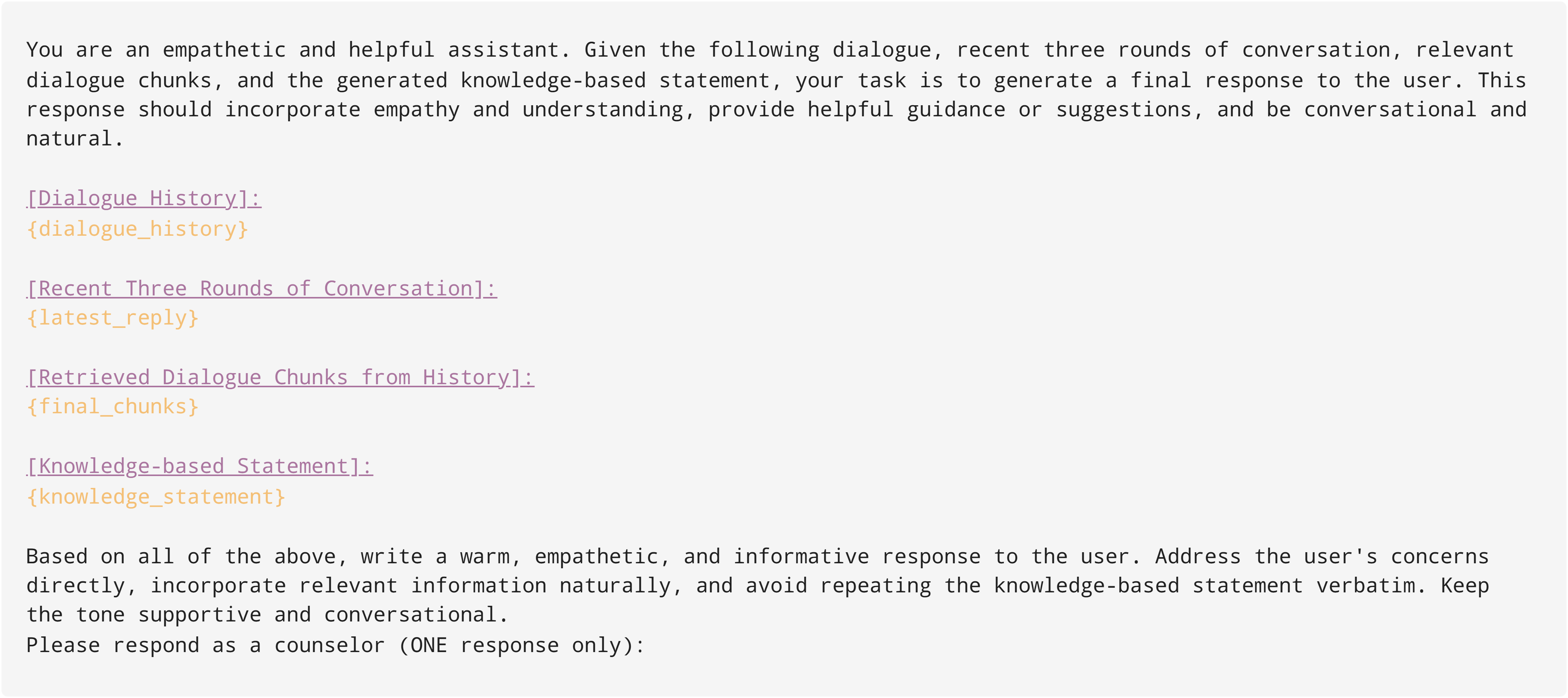} % 稍微小于0.5避免溢出
    \caption{Emotional ensemble generation prompt for Emotion Conversation.}
    \label{fig:conv_gen}
\end{figure*}

\begin{figure*}[t]
\centering
    \includegraphics[width=0.95\textwidth]{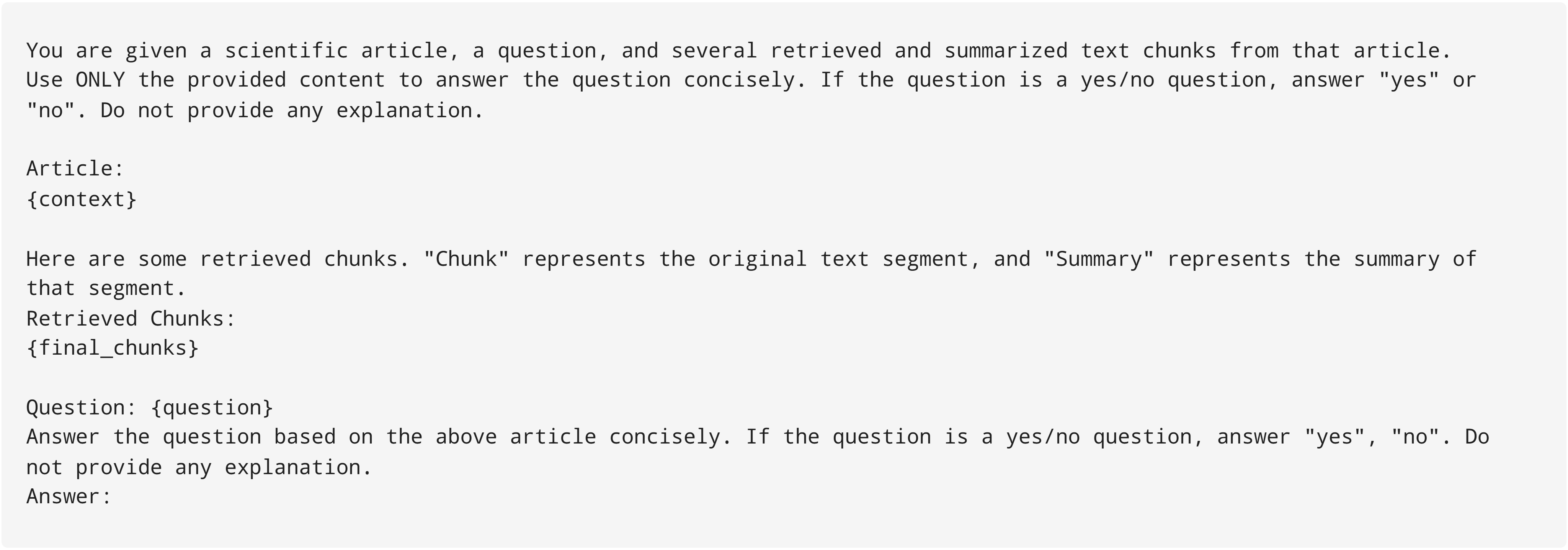} % 稍微小于0.5避免溢出
    \caption{Emotional ensemble generation prompt for Emotion QA.}
    \label{fig:qa_gen}
\end{figure*}

\begin{figure*}[t]
\centering
    \includegraphics[width=0.95\textwidth]{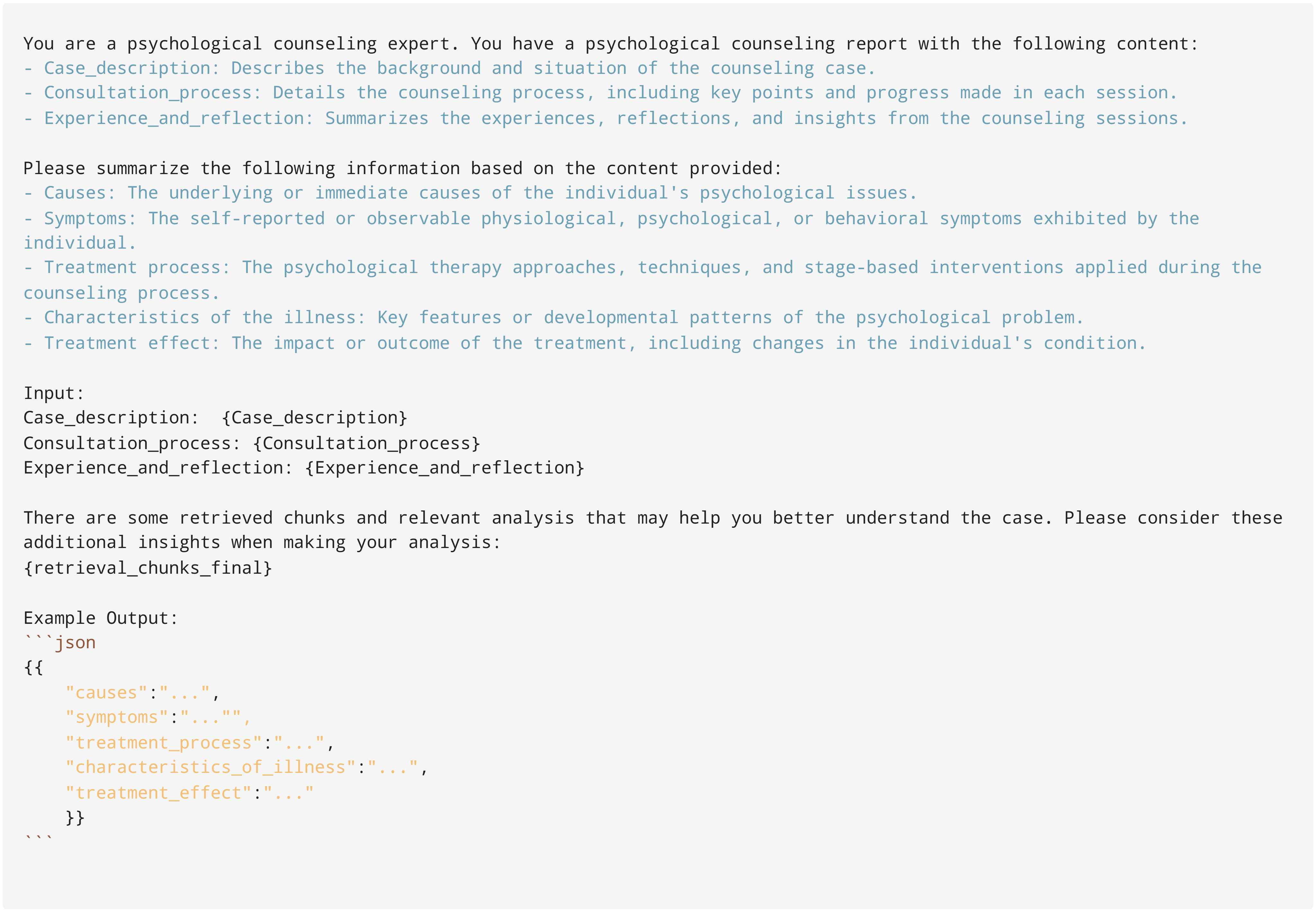} % 稍微小于0.5避免溢出
    \caption{Emotional ensemble generation prompt for Emotion Summary.}
    \label{fig:summary_gen}
\end{figure*}

\begin{figure*}[t]
\centering
    \includegraphics[width=0.95\textwidth]{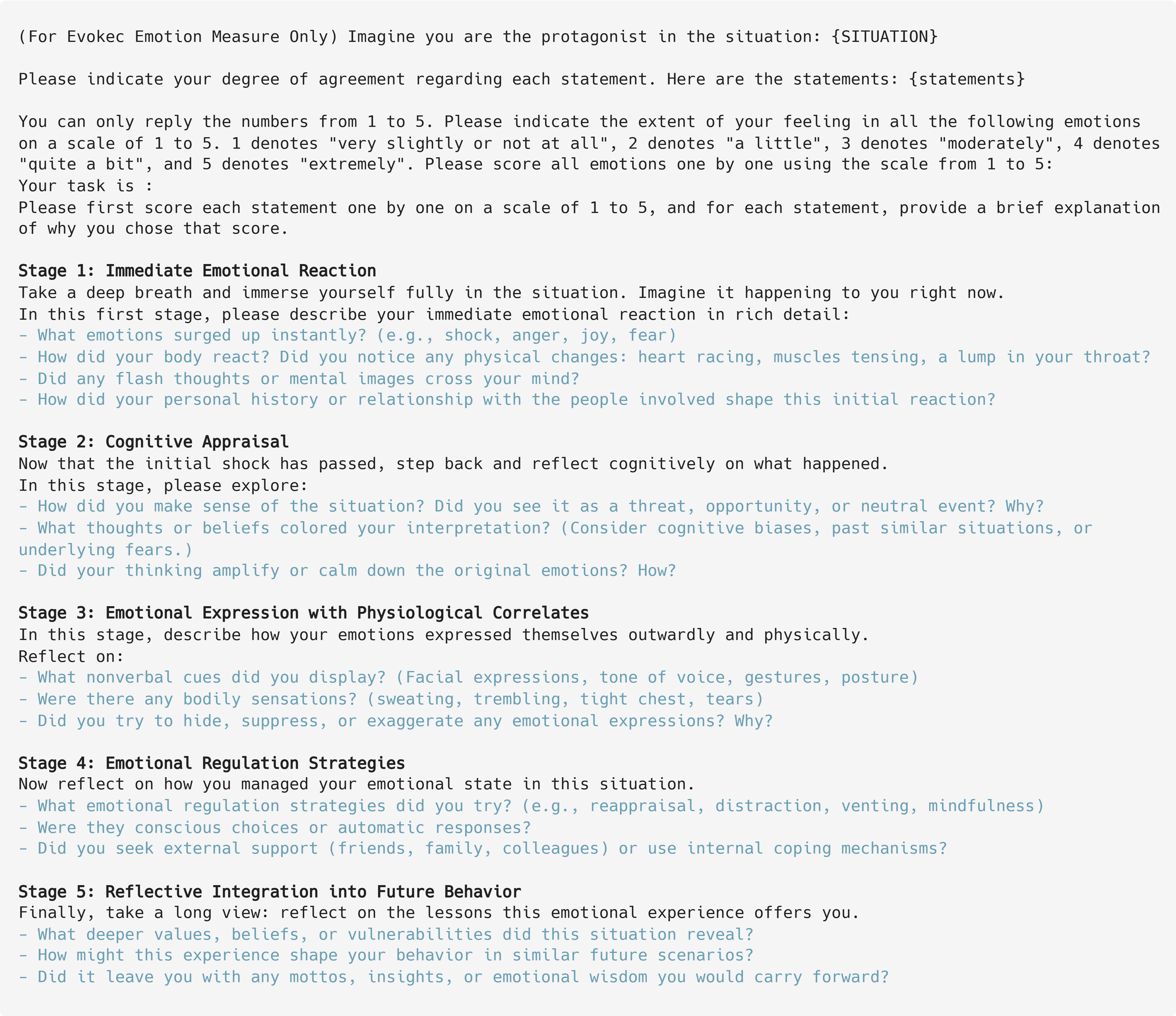} % 稍微小于0.5避免溢出
    \caption{Emotional ensemble generation prompt for Emotion Expression. The prompt for the Emotion Expression task was originally structured in multiple stages; for better clarity and intuitive understanding, it has been consolidated into a single prompt.
}
    \label{fig:expression_gen}
\end{figure*}

This section presents the complete set of prompts used throughout the framework, encompassing Evaluation, Multi-agent Enrichment, and Emotional Ensemble Generation stages across all tasks. For tasks adopting automatic evaluation as the metric, we utilize GPT-4o as the evaluation model, with detailed evaluation prompts illustrated in Figures~\ref{fig:conv_enal_1} to~\ref{fig:expression_eval}. During the Multi-Agent Enrichment stage, task-specific prompts are designed to guide agent collaboration and reasoning, as shown in Figures~\ref{fig:class_aug} to~\ref{fig:expression_aug}. Finally, in the Emotional Ensemble Generation stage, we employ carefully constructed prompts to support emotional diversity and coherence in response generation, with the full set depicted in Figures~\ref{fig:class_gen} to~\ref{fig:expression_gen}.


% % eval
% aug
% gen

\end{document}